\documentclass{article}

\usepackage[preprint]{neurips_2025}

\usepackage[utf8]{inputenc} % allow utf-8 input
\usepackage[T1]{fontenc}    % use 8-bit T1 fonts
\usepackage{hyperref}       % hyperlinks
\usepackage{url}            % simple URL typesetting
\usepackage{booktabs}       % professional-quality tables
\usepackage{amsfonts}       % blackboard math symbols
\usepackage{nicefrac}       % compact symbols for 1/2, etc.
\usepackage{microtype}      % microtypography
\usepackage{xcolor}         % colors
\usepackage{graphicx}
\usepackage{amssymb}
\usepackage{wrapfig}
\usepackage{subfig}
\usepackage{makecell}
\usepackage{multirow}
\usepackage{marvosym}

\usepackage{makecell}
\usepackage{multirow}

\definecolor{nbarrier}{RGB}{255, 120, 50}
\definecolor{nbicycle}{RGB}{255, 192, 203}
\definecolor{nbus}{RGB}{255, 255, 0}
\definecolor{ncar}{RGB}{0, 150, 245}
\definecolor{nconstruct}{RGB}{0, 255, 255}
\definecolor{nmotor}{RGB}{200, 180, 0}
\definecolor{npedestrian}{RGB}{255, 0, 0}
\definecolor{ntraffic}{RGB}{255, 240, 150}
\definecolor{ntrailer}{RGB}{135, 60, 0}
\definecolor{ntruck}{RGB}{160, 32, 240}
\definecolor{ndriveable}{RGB}{255, 0, 255}
\definecolor{nother}{RGB}{139, 137, 137}
\definecolor{nsidewalk}{RGB}{75, 0, 75}
\definecolor{nterrain}{RGB}{150, 240, 80}
\definecolor{nmanmade}{RGB}{213, 213, 213}
\definecolor{nvegetation}{RGB}{0, 175, 0}

\definecolor{nvcolor}{RGB}{119,185,0}
\definecolor{roadcolor}{RGB}{234,51,246}
\definecolor{sidewalkcolor}{RGB}{68,8,72}
\definecolor{parkingcolor}{RGB}{241,156,249}
\definecolor{othergroundcolor}{RGB}{160,32,76}
\definecolor{buildingcolor}{RGB}{246,202,69}
\definecolor{carcolor}{RGB}{111,149,238}
\definecolor{truckcolor}{RGB}{74,32,172}
\definecolor{bicyclecolor}{RGB}{136,227,242}
\definecolor{motorcyclecolor}{RGB}{37,59,146}
\definecolor{othervehiclecolor}{RGB}{96,81,242}
\definecolor{vegetationcolor}{RGB}{79, 173, 50}
\definecolor{trunkcolor}{RGB}{126, 65, 22}
\definecolor{terraincolor}{RGB}{171, 238, 105}
\definecolor{personcolor}{RGB}{234, 60, 49}
\definecolor{bicyclistcolor}{RGB}{234, 66, 195}
\definecolor{motorcyclistcolor}{RGB}{138, 42, 90}
\definecolor{fencecolor}{RGB}{238, 128, 69}
\definecolor{polecolor}{RGB}{252, 241, 161}
\definecolor{trafficsigncolor}{RGB}{233, 51, 35}
\definecolor{other-struct.color}{RGB}{255, 150, 0}
\definecolor{other-objectcolor}{RGB}{50, 255, 255}
\definecolor{lane-markingcolor}{RGB}{150, 255, 170}
\definecolor{color1}{RGB}{176, 36, 24}
\definecolor{color2}{RGB}{0, 176, 80}
\definecolor{color3}{RGB}{0, 0, 200}

\title{QuadricFormer: Scene as Superquadrics for 3D Semantic Occupancy Prediction}

\vspace{-2mm}
\author{
  Sicheng Zuo$^{1,}$\thanks{Equal contribution.  
  \textsuperscript{\textdagger}Project leader.
  } \quad
  Wenzhao Zheng$^{1,*,\dagger}$ \quad 
  Xiaoyong Han$^{1,*}$ \\
  \textbf{Longchao Yang}$^{2}$ \quad
  \textbf{Yong Pan}$^{2}$ \quad
  \textbf{Jiwen Lu}$^{1}$
  \vspace{2mm}
  \\
  $^1$Tsinghua University \quad $^2$Li Auto Inc. \\
  Project Page: \url{https://zuosc19.github.io/QuadricFormer/}
}

\begin{document}

\maketitle

\vspace{-2mm}
\maketitle
\vspace{-8mm}
\begin{center}
    \centering
    \includegraphics[width=\linewidth]{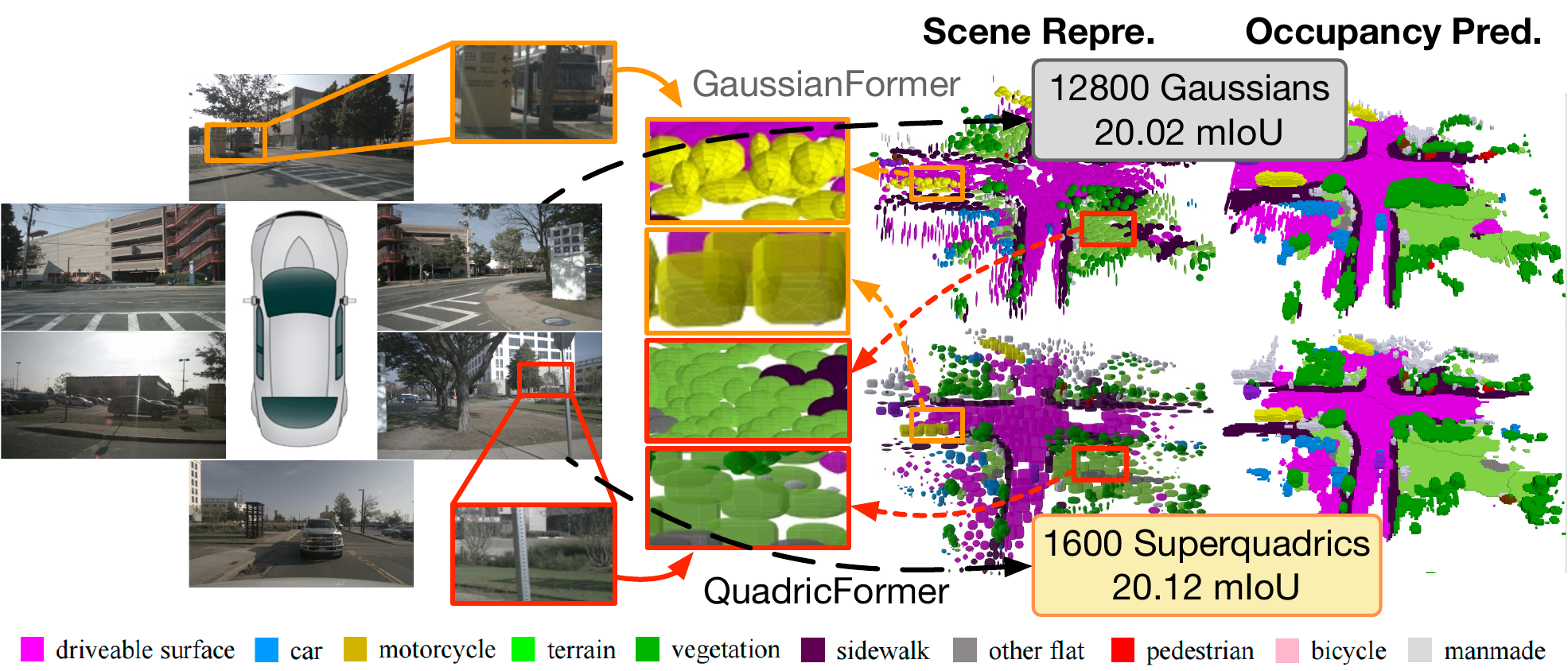}
    \vspace{-4mm}
    \captionof{figure}{Considering the ellipsoidal shape prior of Gaussians, we propose leveraging expressive superquadrics to build an efficient and powerful object-centric representation. Our QuadricFormer achieves state-of-the-art performance with superior efficiency for 3D occupancy prediction.
}
\label{teaser}
\end{center}

\begin{abstract}
3D occupancy prediction is crucial for robust autonomous driving systems as it enables comprehensive perception of environmental structures and semantics.
Most existing methods employ dense voxel-based scene representations, ignoring the sparsity of driving scenes and resulting in inefficiency.
Recent works explore object-centric representations based on sparse Gaussians, but their ellipsoidal shape prior limits the modeling of diverse structures.
In real-world driving scenes, objects exhibit rich geometries (e.g., cuboids, cylinders, and irregular shapes), necessitating excessive ellipsoidal Gaussians densely packed for accurate modeling, which leads to inefficient representations.
% To address this, we propose to use superquadrics—geometrically expressive primitives—as an alternative to Gaussians, which can efficiently represent complex structures without much overlap.
To address this, we propose to use geometrically expressive superquadrics as scene primitives, enabling efficient representation of complex structures with fewer primitives through their inherent shape diversity.
We develop a probabilistic superquadric mixture model, which interprets each superquadric as an occupancy probability distribution with a corresponding geometry prior, and calculates semantics through probabilistic mixture.
Building on this, we present QuadricFormer, a superquadric-based model for efficient 3D occupancy prediction, and introduce a pruning-and-splitting module to further enhance modeling efficiency by concentrating superquadrics in occupied regions.
% We then employ a superquadric-based model (QuadricFormer) for efficient 3D occupancy prediction, and design a pruning-and-splitting module to further improve modeling efficiency by concentrating superquadrics in occupied regions.
Extensive experiments on the nuScenes dataset demonstrate that QuadricFormer achieves state-of-the-art performance while maintaining superior efficiency.
Code is available at \href{https://github.com/zuosc19/QuadricFormer}{{\texttt{https://github.com/zuosc19/QuadricFormer}}}.

\end{abstract}    
\section{Introduction}
Vision-centric autonomous driving systems have gained much attention for their cost-effectiveness over LiDAR-based solutions~\cite{monoscene,tpvformer,occformer,fb-occ,voxformer}. 
However, they struggle to perceive irregularly shaped obstacles due to visual ambiguity, which compromises driving safety.
Recent advances in 3D semantic occupancy prediction address this by estimating voxel-level occupancy status and semantic labels in 3D scenes~\cite{openoccupancy,surroundocc,tian2023occ3d,scene_as_occ}. 
This provides a full understanding of scene structures and semantics, which enables applications including self-supervised 3D scene understanding~\cite{cao2023scenerf,selfocc}, 4D occupancy forecasting~\cite{occworld,driveworld,wang2024occsora,yan2024renderworld}, and end-to-end autonomous driving~\cite{hu2023uniAD,zheng2024gaussianad}.

Despite promising applications, 3D semantic occupancy prediction faces efficiency challenges due to its dense 3D predictions~\cite{monoscene,tian2023occ3d}.
An efficient and expressive 3D representation is therefore essential.
While voxel-based methods~\cite{voxformer, surroundocc} use dense 3D grids to capture fine details, they ignore the sparsity of driving scenes and suffer from high computational costs.
Recent advances introduce object-centric representations using 3D Gaussians~\cite{gaussianformer, zuo2024gaussianworld} to describe scenes sparsely.
Each Gaussian models the occupancy probability distribution of its local region via learnable attributes including position, covariance, opacity, and semantics.
However, Gaussian representations are fundamentally limited.
By their mathematical formulation, Gaussians describe the spatial occupancy probability with an ellipsoidal decay pattern. 
This imposes a strong ellipsoidal shape prior to Gaussians and severely constrains their capacity to model diverse geometries.
Real-world driving scenarios contain objects with rich structural variations, which cannot be accurately represented by a few ellipsoidal Gaussian. 
Consequently, Gaussian-based models must aggregate numerous densely packed Gaussians to approximate target shapes, causing significant efficiency degradation.

In this paper, we propose an efficient and expressive object-centric 3D representation using superquadrics~\cite{barr1981superquadrics} as scene primitives.
Superquadrics are a family of parameterized shapes with high geometric expressiveness and compact shape parameters, offering great flexibility in modeling diverse geometries.
This allows superquadrics to model complex structures with sparse packing, enabling an efficient and powerful representation~\cite{fedele2025superdec}.
We represent scenes with a set of learnable superquadrics, each characterized by attributes including position, scale, rotation, opacity, semantics, and shape exponents.
For occupancy prediction, we adopt a probabilistic superquadric mixture model that interprets each superquadric as a local occupancy probability distribution, and calculates semantics through probabilistic mixture.
Building on this representation, we introduce QuadricFormer, a superquadric-based framework for efficient 3D semantic occupancy prediction. Moreover, we design a pruning-and-splitting module that concentrates superquadrics on occupied regions to further enhance modeling efficiency. 
Extensive experiments on the nuScenes dataset demonstrate that our QuadricFormer achieves state-of-the-art performance with superior efficiency.
\section{Related Work}
\subsection{3D Semantic Occupancy Prediction}
3D semantic occupancy prediction reconstructs fine-grained 3D scenes by labeling each voxel with geometric and semantic information, which is critical for autonomous driving~\cite{monoscene,tpvformer,tian2023occ3d,occformer,occworld}.
LiDAR and cameras are the two most commonly used sensors.
While LiDAR-based methods excel in depth accuracy~\cite{MV3D-LiDAR,dsketch,af2s3net,pointpillars,aicnet,amvnet,lmscnet,spvnas,js3c,ye2022lidarmultinet,drinet++,voxelnet,pointocc}, their limitations in adverse weather and long-range detection motivate the vision-centric approaches, which reconstruct scenes from multi-view visual input~\cite{fb-occ,surroundocc,yu2023flashocc,monoscene,tpvformer}.
Early approaches lifted image features directly into dense voxel grids for 3D occupancy prediction\cite{chen20193d,voxformer,surroundocc,cotr}. 
However, given the sparsity of occupied voxels in driving scenes, subsequent works prioritized efficiency through alternative representations.
Planar representations like BEV\cite{bevformer} and TPV\cite{tpvformer} compress 3D data into 2D feature maps for efficient processing, but sacrifice geometric fidelity.
Object-centric modeling preserves geometric fidelity by focusing computation on salient regions\cite{gaussianformer,huang2024probabilistic,octreeocc,sparseocc,zuo2024gaussianworld}, alleviating both the redundancy of uniform voxel grids and the information loss from planar compression.
However, these methods still struggle to balance efficiency and modeling capacity due to the complexity of real-world structures.
To address this, we propose a superquadric-based model that achieves efficient and accurate representation of complex geometries.

\subsection{Object-centric scene representations}
Existing 3D scene representations primarily use voxel-based frameworks for fine-grained volumetric modeling~\cite{surroundocc,voxformer}, excelling in semantic prediction tasks. 
However, their uniform processing of all voxels introduces spatial redundancy, particularly in sparse environments.
To address this, recent works explore object-centric representations~\cite{sparseocc,octreeocc,gaussianformer,huang2024probabilistic,zuo2024gaussianworld}.
One line of methods partitions dense grids into localized regions, preserving only detected object areas~\cite{sparseocc,octreeocc}.
While efficient, non-empty regions may be falsely pruned, leading to irreversible loss of critical geometry.
Alternatively, point-based methods use sparse points as queries for iterative refinement~\cite{occaspoints,wang2024opus}.
However, points inherently lack spatial extent, limiting their ability to capture contextual geometry. 
Recent advances adopt 3D semantic Gaussians~\cite{gaussianformer,huang2024probabilistic,zuo2024gaussianworld}, where probability densities radiate from Gaussian centers to enable adaptive spatial coverage.
While Gaussians mitigate point rigidity through probability spread, complex geometries often require multiple densely packed primitives, particularly for fine structures, leading to inefficient representations.
In this paper, we propose geometrically expressive superquadrics as compact scene primitives. Unlike conventional object-centric methods, superquadrics natively parameterize diverse geometries (e.g., cuboids, cylinders) without dense packing, achieving superior reconstruction fidelity using fewer primitives.

\subsection{Superquarics}
Superquadrics are parametric geometric primitives introduced by Barr et al.~\cite{barr1981superquadrics} to model diverse shapes with compact parameterizations. 
A canonical superquadric is defined by five parameters: three scale parameters along each of its semi-axes and two exponents that determine its shape~\cite{jaklic2000segmentation}. 
The scale and shape parameters of superquadrics allow for smooth interpolation between different geometric shapes, such as cuboids, cylinders, and spheres. 
When combined with six pose parameters for translation and rotation, a superquadric can represent a complete 3D object using only 11 parameters.
Recent works employed superquadrics to decompose complex environments into compact geometric primitives~\cite{fedele2025superdec}.
These methods demonstrate compelling reconstruction capability and editing flexibility, while maintaining model efficiency.
However, existing approaches operate exclusively on point clouds and are limited to object-level reconstructions.
Differently, we present the first superquadric-based framework for holistic scene reconstruction directly from multi-view images, delivering state-of-the-art performance with superior efficiency.

\begin{figure*}[t]
    \centering
    \subfloat[\label{fig:repre}Scene Representations.]{
        \includegraphics[width=0.52\linewidth]{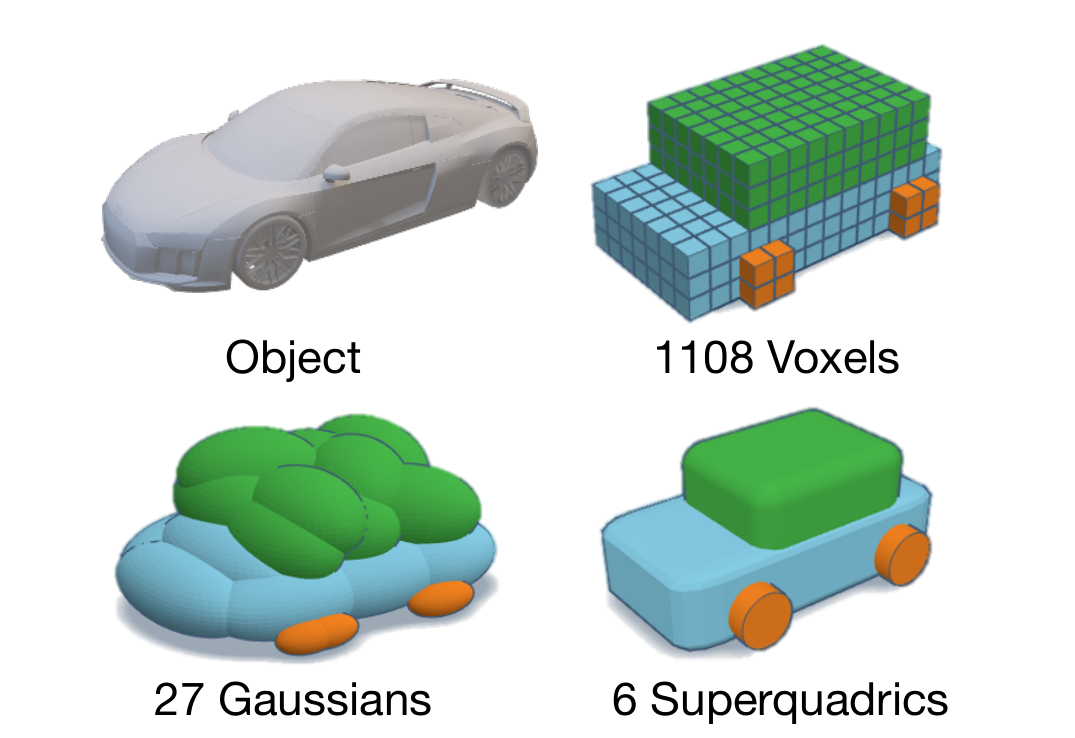}
    }
    \subfloat[\label{fig:performance}Performance Comparison.]{
        \includegraphics[width=0.46\linewidth]{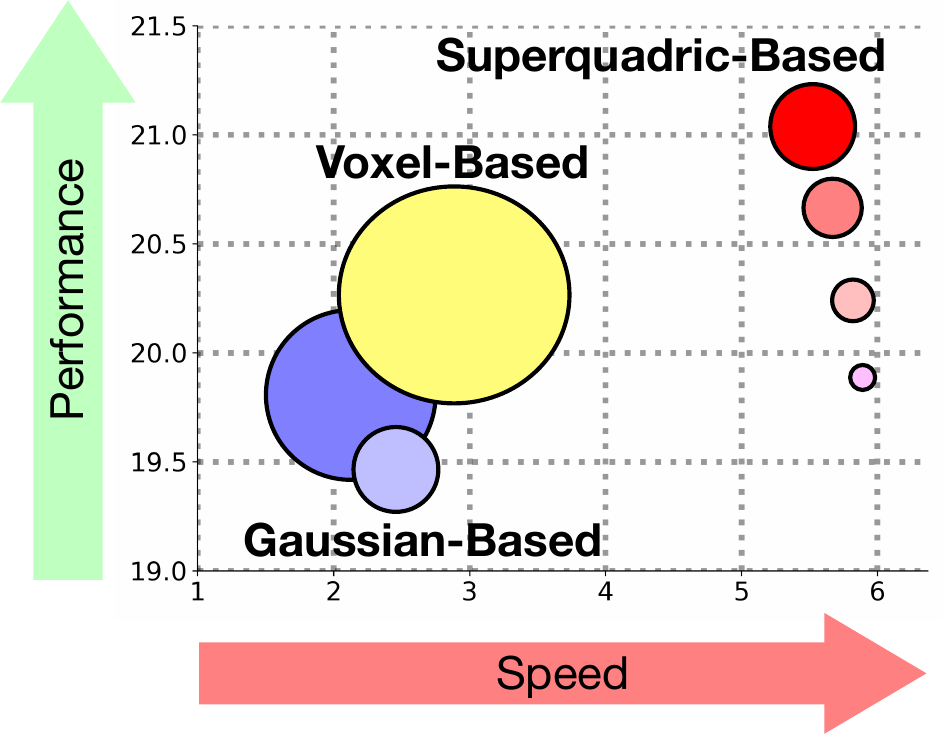}
    }
    \caption{\textbf{Comparisons between different representations.} (a) Quadric-based method represents the same object with a smaller number of primitives and greater shape expressiveness. (b) Quadric-based representation outperforms existing methods in both accuracy and speed with far fewer primitives.}
    \vspace{-3mm}
\end{figure*}
\section{Proposed Approach}
In this section, we present our method based on the superquadric representation for efficient 3D semantic occupancy prediction.
We first review the Gaussian-based object-centric representation and analyze its limitations (Sec~\ref{sec-3.1}).
We then introduce our superquadric representation and probabilistic modeling approach for efficient occupancy prediction (Sec~\ref{sec-3.2}).
Finally, we describe the overall architecture of QuadricFormer for vision-centric 3D occupancy prediction.(Sec~\ref{sec-3.3}).

\subsection{Object-Centric Representation} \label{sec-3.1}
Vision-centric 3D semantic occupancy prediction aims to estimate the occupancy status and semantic label of each voxel in 3D space based on visual inputs.
Formally, given input images $\mathcal{I}=\{\mathbf{I}_i\}_{i=1}^{N}$ from $N$ views, the model aims to predict voxel-level semantic labels $\mathbf{O}\in \mathcal{C}^{X \times Y \times Z}$ of the 3D scene, where $\mathcal{C}$ denotes the semantic classes and $X \times Y \times Z$ represents the spatial shape of occupancy.

To achieve this, voxel-based methods~\cite{surroundocc,occformer} adopt dense voxel features to model 3D scenes, resulting in extremely high computational complexity of $\mathcal{O}(XYZ)$.
This inefficiency stems from their uniform processing of all voxels in space, which ignores the inherent sparsity of real-world scenes.
Considering this, recent works~\cite{gaussianformer, huang2024probabilistic} explore object-centric representations based on 3D Gaussians to focus computational resources on salient regions for efficient scene modeling.
Gaussian-based method~\cite{huang2024probabilistic} typically employs a set of $P$ semantic 3D Gaussian primitives $\mathcal{G}=\{\mathbf{G}_i\}_{i=1}^{P}$ to represent 3D scenes sparsely.
Each Gaussian $\mathbf{G}_i$ models a flexible local region with its explicit mean $\mathbf{m}_i$, scale $\mathbf{s}_i$, rotation $\mathbf{r}_i$, opacity $a_i$, and semantic probability $\mathbf{c}_i$.
% To obtain geometric occupancy prediction, GaussianFormer-2~\cite{gaussianformer-2} interprets each Gaussian as an occupancy probability distribution of its local region.
For a point $\mathbf{x}$ in 3D space, its geometric occupancy probability associated with the Gaussian $\mathbf{G}$ is:
\vspace{-1mm}
\begin{equation}
    \alpha(\mathbf{x};\mathbf{G}) = {\rm{exp}}\big(-\frac{1}{2}(\mathbf{x}-\mathbf{m})^{\rm T} \mathbf{\Sigma}^{-1} (\mathbf{x}-\mathbf{m})\big),
    \label{eq: single prob}
\end{equation}
\vspace{-2mm}
\begin{equation}
    \mathbf{\Sigma} = \mathbf{R}\mathbf{S}\mathbf{S}^T\mathbf{R}^T, \quad \mathbf{S} = {\rm{diag}}(\mathbf{s}), \quad \mathbf{R} = {\rm{q2r}}(\mathbf{r}),
\end{equation}
% \vspace{-1mm}
where $\mathbf{x}$ denotes the point position, and $\mathbf{\Sigma}$, $\mathbf{R}$, $\mathbf{S}$ represent the covariance matrix, the rotation matrix constructed from the quaternion $\mathbf{r}$, and the diagonal scale matrix from the scale $\mathbf{s}$.
Furthermore, a probabilistic Gaussian mixture model is used to aggregate multiple Gaussians for predicting the structure and semantics of the scene. 
As each Gaussian represents a flexible region of the scene, the Gaussian-based representation enables adaptive allocation of resources and efficient modeling.
% To derive the overall probability of occupancy, it is assumed that the probabilities of a point being occupied by different Gaussians are independent.
% Then the final occupancy probability can be computed as:
% \vspace{-2mm}
% \begin{equation}
%     \alpha(\mathbf{x}) = 1 - \prod_{i=1}^{P}\big(1 - \alpha(\mathbf{x};\mathbf{G}_i)\big).
%     \label{eq: multi prob}
% \end{equation}
% \vspace{-2mm}
% For semantic prediction, the semantic probabilities of different Gaussians are aggregated:
% % Besides geometry prediction, the semantic probabilities of different Gaussians are aggregated to compute semantic probabilities:
% \vspace{-0mm}
% \begin{equation}
%     \mathbf{e}(\mathbf{x};\mathcal{G}) = \frac{\sum_{i=1}^{P}p(\mathbf{x}|\mathbf{G}_i)a_i{\mathbf{c}}_i}{\sum_{j=1}^{P}p(\mathbf{x}|\mathbf{G}_j)a_j},
% \label{eq: gmm}
% \end{equation}
% \vspace{-2mm}
% \begin{small}
% \begin{equation}
%     p(\mathbf{x}|\mathbf{G}_i) = \frac{1}{(2\pi)^{\frac{3}{2}}|\mathbf{\Sigma}|^{\frac{1}{2}}}{\rm{exp}}\big(-\frac{1}{2}(\mathbf{x}-\mathbf{m})^{\rm T} \mathbf{\Sigma}^{-1} (\mathbf{x}-\mathbf{m})\big),
% \end{equation}
% \end{small}where the weight of each Gaussian is normalized based on their contributions to the point. 

Although 3D Gaussian representation is more efficient than dense voxels (e.g., $6400$ Gaussians vs. $200\times200\times16$ voxels per scene), it still exhibits limitations that prevent an optimal efficiency-performance balance.
\textit{Our key insight is that Gaussians inherently impose an ellipsoidal shape prior, which limits their ability to model diverse structures}.
As shown in Eq.~\ref{eq: single prob}, the occupancy probability distribution of the Gaussian $\mathbf{G}$ can be viewed as a set of iso-probability surfaces defined by:
% $g(\mathbf{x};\mathbf{G})=-\frac{1}{2}(\mathbf{x}-\mathbf{m})^{\rm T} \mathbf{\Sigma}^{-1} (\mathbf{x}-\mathbf{m})=k$, where the occupancy probability $\alpha(\mathbf{x};\mathbf{G})$ decreases as $k$ increases.
% For simplicity, we omit the rotation and mean of the Gaussian, then the iso-probability surfaces of a Gaussian can be simplified to:
\vspace{-1mm}
\begin{equation}
    g(\mathbf{x})=-\frac{1}{2}\big((\frac{x}{s_x})^{2}+(\frac{y}{s_y})^{2}+(\frac{z}{s_z})^{2}\big)=k,
\label{eq: ellipsoid}
\end{equation}
\vspace{-0mm}where $\mathbf{x}=\left(x, y, z\right)^T$ denotes the point position, $k$ denotes the hyperparameter of the surface family, and $\mathbf{s}=\left(s_x, s_y, s_z\right)^T$ represents the Gaussian's scales along three axes.
The rotation and mean of the Gaussian are omitted for simplicity in Eq.~\ref{eq: ellipsoid}, which describes a standard ellipsoid.
Each Gaussian then models occupancy probability with an ellipsoidal decay in 3D space.
But real-world objects often have diverse shapes, such as cuboids, cylinders, and irregular shapes, which cannot be accurately represented by a few ellipsoidal Gaussians.
This forces the model to use numerous densely packed Gaussians to approximate complex structures, leading to inefficient scene representations. 
% Moreover, under the probabilistic modeling mechanism as in Eq.~\ref{eq: multi prob}, the overall occupancy probability increases as more Gaussians overlap.
% This further encourages unnecessary overlapping of Gaussians, diminishing the modeling efficiency. 
% These limitations arise from the inherent ellipsoidal shape prior in Gaussian modeling. 
In contrast, our method employs expressive superquadrics as scene primitives, enabling efficient and compact modeling of complex structures with only a few sparsely packed superquadrics.

% Thus, each Gaussian inherently models occupancy probabilities with an ellipsoidal decay pattern in 3D space.
% However, objects in real-world driving scenarios exhibit highly diverse structures, such as planes, cubes, cylinders, and their complex combinations, which cannot be accurately represented by a single ellipsoidal Gaussian.
% As a result, the model is forced to aggregate a large number of overlapping Gaussians to approximate complex object structures, significantly reducing the representation efficiency.
% Furthermore, under the probabilistic modeling mechanism as in Eq.~\ref{eq: multi prob}, the overall occupancy probability increases as more Gaussians overlap. During optimization, this encourages the model to further cluster Gaussians in occupied regions to boost prediction probabilities, further diminishing the modeling efficiency of the 3D Gaussian representation.
% These limitations stem from the inherent ellipsoidal shape prior in Gaussian probabilistic modeling, which undermines the effectiveness and efficiency of 3D Gaussian representation.
% Our method employs geometrically expressive superquadrics as scene primitives, allowing for efficient modeling of complex structures without much overlapping and enabling a more efficient and powerful object-centric representation.

\begin{figure*}[t]
\centering
\includegraphics[width=1.0\textwidth]{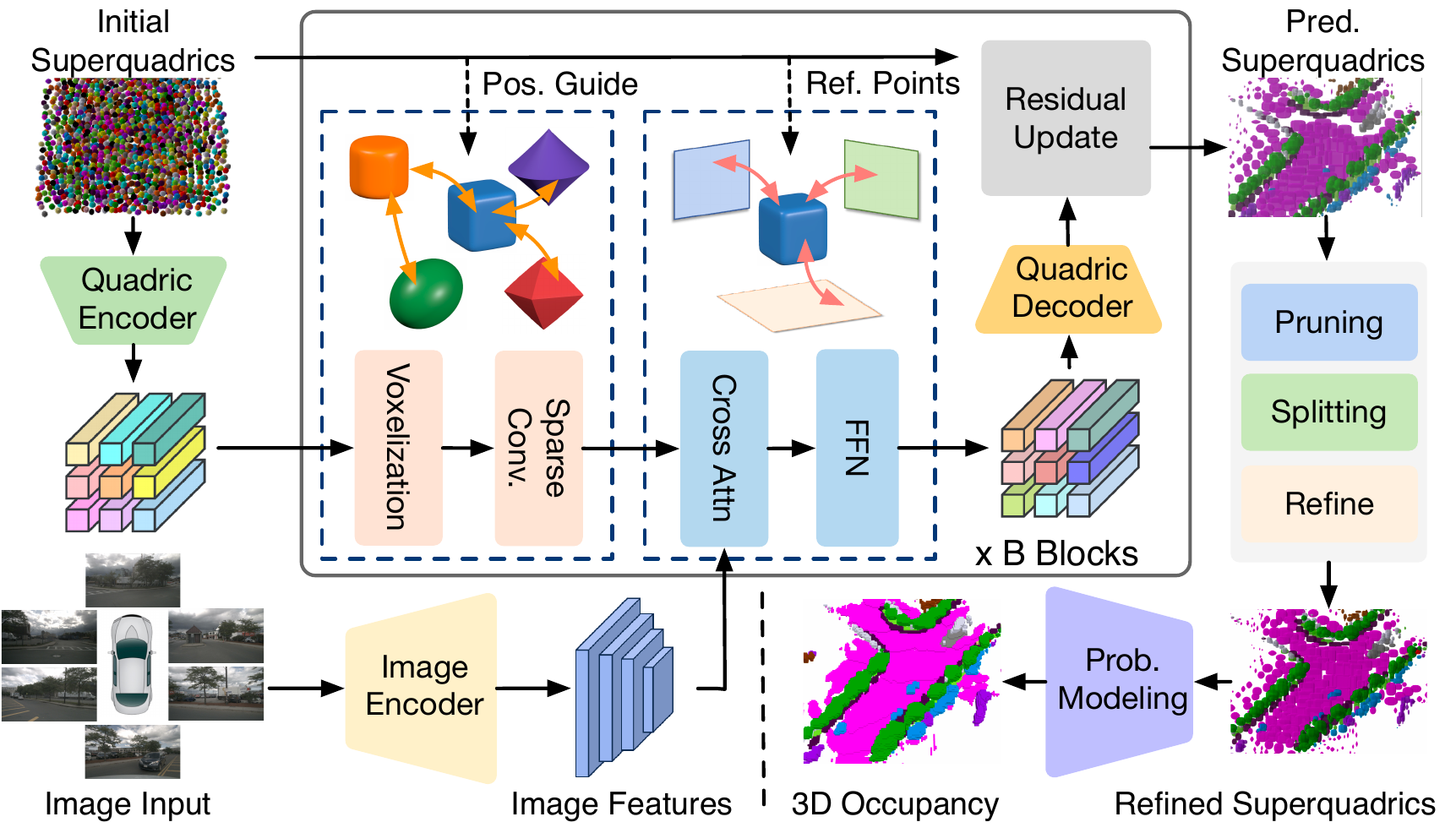}
\vspace{-5mm}
\caption{\textbf{Overall Framework of QuadricFormer.}We use several quadric-encoder blocks to update superquadrics, and employ a pruning-and-splitting module to further enhance modeling efficiency.
}
\label{fig:framework}
\vspace{-5mm}
\end{figure*}

\subsection{Scene as Superquadrics} \label{sec-3.2}
We introduce an object-centric scene representation leveraging superquadric primitives for their efficiency and expressive power.
Superquadrics are a parametric shape family with strong geometric expressiveness, defined as follows:
\vspace{-2mm}
\begin{equation}\label{eq: superquadric}
    f(\mathbf{x})=\left(\left(\frac{x}{s_{x}}\right)^{\frac{2}{\epsilon_{2}}}+\left(\frac{y}{s_{y}}\right)^{\frac{2}{\epsilon_{2}}}\right)^{\frac{\epsilon_{2}}{\epsilon_{1}}}+\left(\frac{z}{s_{z}}\right)^{\frac{2}{\epsilon_{2}}}=k,
\end{equation}
\vspace{0mm}where $\mathbf{x}=\left(x, y, z\right)^T$ denotes the point position, and $k$ denotes the hyperparameter of the surface family.
Compared to the ellipsoids in Eq.~\ref{eq: ellipsoid}, superquadrics introduce only two additional shape-defining exponents $\epsilon_{1}, \epsilon_{2}$ yet can represent a much wider variety of shapes.
As shown in Fig.~\ref{fig:repre}, superquadrics allow for continuous and diverse shape variations as the shape parameters change.
% superquadrics can smoothly represent cuboids, cylinders, ellipsoids, pyramids, and other complex structures. 
% This inherent parameter efficiency and geometric expressiveness make superquadrics well-suited for representing complex 3D structures.
This inherent parameter efficiency and geometric expressiveness enable superquadrics to model diverse shapes without being densely packed. 
Consequently, only a small number of superquadrics are needed to represent complex scene structures, achieving an efficient yet powerful scene representation.

We thus utilize a set of $P$ parameterized superquadrics $\mathcal{Q}=\{\mathbf{Q}_i\}_{i=1}^{P}$ to represent the 3D scene.
Each superquadric is characterized by its scale $\mathbf{s}$ and shape exponents $\epsilon_{1}, \epsilon_{2}$ to define its geometry.
To extend the representation to the global coordinate system, each primitive is also assigned a position $\mathbf{x}$ and rotation $\mathbf{r}$.
Beyond geometric attributes, each superquadric is further equipped with an opacity $a$ and a semantic probability $\mathbf{c}$ to incorporate semantic information.
In summary, our superquadric-based representation can be formulated as:
\vspace{-0mm}
\begin{equation}
    \mathcal{Q}=\{\mathbf{Q}_i\}_{i=1}^{P}=\{(\mathbf{x}_i, \mathbf{s}_i, \mathbf{r}_i, a_i, \epsilon_{1;i}, \epsilon_{2;i}, \mathbf{c}_i,)\}_{i=1}^{P}.
\end{equation}\vspace{-5mm}

We now explore obtaining 3D occupancy prediction from the superquadric representation. 
Existing methods~\cite{fedele2025superdec} typically treat superquadrics as deterministic surfaces, fitting them to object parts for point cloud reconstruction. 
However, these surface-based approaches face key limitations in vision-centric occupancy prediction.
A primary challenge is supervision. 
While point cloud reconstruction directly optimizes the distance between points and the superquadric surfaces, occupancy prediction requires fine-grained scene understanding, which lacks clear surface-based constraints. 
Furthermore, surface-based methods rely on the explicit structure from point cloud inputs, whereas visual inputs introduce structural uncertainty, making deterministic modeling unstable. 
Lastly, surface-based methods focus on object-level reconstruction with simple spatial relationships.
But real-world driving scenes involve far more complex surface interactions, posing significant modeling difficulties.

% We now discuss obtaining 3D semantic occupancy prediction from the superquadric representation.
% Existing methods typically treat each superquadric as a deterministic surface to model object shapes.
% By fitting differently shaped superquadrics to distinct object parts, these approaches achieve compact 3D representations and accurate point cloud reconstruction.
% However, these surface-based methods face several limitations when applied to vision-centric 3D occupancy prediction.
% A primary challenge is supervision.
% In point cloud reconstruction, the model can be directly optimized by minimizing the distance between ground-truth points and the superquadric surfaces.
% In contrast, 3D semantic occupancy prediction, which requires inferring fine-grained scene structure and semantics, lacks straightforward surface-based constraints for superquadrics. 
% Furthermore, surface-based methods generally rely on point cloud inputs, where the structural information is explicit and well-suited for deterministic surface modeling.
% However, vision-based methods suffer from inherent uncertainty in 3D structure, and deterministic modeling can lead to unstable optimization or even training collapse.
% Lastly, surface-based methods typically focus on object-level reconstruction with simple spatial relationships.
% In real-world driving scenes, the interactions between object surfaces are far more complex, making surface-based modeling highly challenging.

To achieve robust 3D semantic occupancy prediction, we design a probabilistic modeling mechanism that converts superquadrics into occupancy probabilities.
Inspired by GaussianFormer-2~\cite{huang2024probabilistic}, we adopt a probabilistic superquadric mixture model, where each superquadric defines the occupancy probability distribution in its local neighborhood.
To compute the probability of a 3D point $\mathbf{x}$ being occupied by the superquadric $\mathbf{Q}$, we first transform $\mathbf{x}$ into $\mathbf{Q}$'s local coordinate system, defined by its position $\mathbf{m}$ and rotation 
\vspace{-2mm}
$\mathbf{r}$:
\begin{equation}
    \mathbf{x}_{\mathbf{Q}} = \mathbf{R}(\mathbf{x}-\mathbf{m}),
    \label{eq: coord transform}
\end{equation}\vspace{-0mm}where $\mathbf{x}_{\mathbf{Q}}$ denotes the local coordinate of $\mathbf{x}$, and $\mathbf{R}$ denotes the rotation matrix constructed from the rotation $\mathbf{r}$.
The occupancy probability of $\mathbf{x}$ associated with $\mathbf{Q}$ is then computed as:
\vspace{-2mm}
\begin{equation}
    p_{\mathbf{o}}(\mathbf{x};\mathbf{G}) = \rm{exp}\big(-f(\mathbf{x}_{\mathbf{Q}})\big) = \rm{exp}\left(\left(\left(\frac{x_{\mathbf{Q}}}{s_{x}}\right)^{\frac{2}{\epsilon_{2}}}+\left(\frac{y_{\mathbf{Q}}}{s_{y}}\right)^{\frac{2}{\epsilon_{2}}}\right)^{\frac{\epsilon_{2}}{\epsilon_{1}}}+\left(\frac{z_{\mathbf{Q}}}{s_{z}}\right)^{\frac{2}{\epsilon_{2}}}\right),
    \label{eq: single prob q}
\end{equation}
where $\mathbf{x}_{\mathbf{Q}}=\left( x_{\mathbf{Q}}, y_{\mathbf{Q}}, z_{\mathbf{Q}} \right)^T$ and $\mathbf{s}=\left( s_x, y_x, z_s\right)^T$ are the position and scale parameters, respectively, and $\epsilon_{1}, \epsilon_{2}$ are the shape exponents of the superquadric $\mathbf{Q}$.
Assuming conditional independence of occupancy among different superquadrics, the final occupancy probability at $\mathbf{x}$ is computed as:
\vspace{-0mm}
\begin{equation}
    p_{\mathbf{o}}(\mathbf{x}) = 1 - \prod_{i=1}^{P}\big(1 - p_{\mathbf{o}}(\mathbf{x};\mathbf{Q}_i)\big).
    \label{eq: multi prob q}
\end{equation}
Semantic predictions are subsequently inferred by a weighted aggregation of semantic probabilities from all contributing superquadrics, where weights correspond to their occupancy influence at $\mathbf{x}$:
\begin{equation}
    p_{\mathbf{c}}(\mathbf{x}) = \frac{\sum_{i=1}^{P}p_{\mathbf{o}}(\mathbf{x}|\mathbf{Q}_i)a_i{\mathbf{c}}_i}{\sum_{j=1}^{P}p_{\mathbf{o}}(\mathbf{x}|\mathbf{Q}_j)a_j},
\label{eq: sem prob q}
\end{equation}
The key to this probabilistic modeling is incorporating the superquadric geometry as shape priors within the probability distribution, realized as iso-probability surfaces conforming to its geometry in Eq.~\ref{eq: superquadric}.
Leveraging the geometrically expressive power of superquadrics, our model can efficiently represent complex 3D structures using a sparse set of primitives without dense packing, achieving an efficient yet powerful scene representation.
Moreover, this probabilistic framework effectively models structural uncertainties arising from visual ambiguities, significantly improving the model's robustness and generalization capabilities.

% \begin{figure*}[t]
% \centering
% \includegraphics[width=1.0\textwidth]{figures/framework.pdf}
% \vspace{-7mm}
% \caption{\textbf{Overall Framework of QuadricFormer.}We use several quadric-encoder blocks to update superquadrics, and employ a pruning-and-splitting module to further enhance modeling efficiency.
% }
% \label{fig:framework}
% \vspace{-5mm}
% \end{figure*}

\subsection{QuadricFormer} \label{sec-3.3}
We present the overall framework of QuadricFormer in Fig.~\ref{fig:framework}.
Starting from the image inputs of $N$ views $\mathcal{I}=\{\mathbf{I}_i\}_{i=1}^{N}$, we first employ an image backbone $E_{\mathbf{I}}$ to extract multi-scale image features $\mathbf{F}_{\mathbf{I}}$:
\vspace{-0mm}
\begin{equation}
     \mathbf{F}_{\mathbf{I}} = E_{\mathbf{I}}(\mathcal{I}),
    \label{eq: img feat}
\end{equation}
Due to the lack of any structural prior of the scene, we randomly initialize a few superquadrics $\mathbf{Q}_{init}$ in 3D space, and use $B$ quadric-encoder blocks $E_{\mathbf{B}}$ to predict the final superquadrics from images.
% \vspace{-0mm}
% \begin{equation}
%      \mathbf{Q} = E_{\mathbf{B}}(\mathbf{Q}_{init}, \mathbf{F}).
%     \label{eq: encode block}
% \end{equation}
In each block, we first encode current superquadrics $\mathbf{Q}_{i}$ into features $\mathbf{F}_{\mathbf{Q}}$ via a quadric encoder $E_{\mathbf{Q}}$:
\vspace{-0mm}
\begin{equation}
     \mathbf{F}_{\mathbf{Q}} = E_{\mathbf{Q}}(\mathbf{Q}_{i}).
    \label{eq: encode quadric}
\end{equation}
We then use 3D sparse convolution $E_{conv}$ for superquadric feature self-encoding and deformable attention $E_{attn}$ for interaction between superquadric and image features:
\vspace{-0mm}
\begin{equation}
     \mathbf{F}_{\mathbf{Q}} = E_{conv}(\mathbf{F}_{\mathbf{Q}}, \mathbf{x}_{\mathbf{Q}}), \mathbf{F}_{\mathbf{Q}} = E_{attn}(\mathbf{F}_{\mathbf{Q}}, \mathbf{x}_{\mathbf{Q}}, \mathbf{F}_{\mathbf{I}}), 
    \label{eq: encode quadric feat}
\end{equation}
where $\mathbf{x}_{\mathbf{Q}}$ denotes the explicit position of the superquadric ${\mathbf{Q}}$, serving as auxiliary information to guide feature encoding.
Finally, a quadric decoder $D_{\mathbf{Q}}$ is used to predict the update of superquadric attributes $\Delta \mathbf{Q}$, which are combined with the original attributes $\mathbf{Q}$ via residual addition:
\vspace{-0mm}
\begin{equation}
     \Delta \mathbf{Q} = D_{\mathbf{Q}}(\mathbf{F}_{\mathbf{Q}}), \mathbf{Q}_{i+1} = \mathbf{Q}_{i} + \Delta \mathbf{Q}.
    \label{eq: decode quadric}
\end{equation}
After $B$ blocks update, we get the final superquadric prediction $\mathbf{Q}$, and the 3D semantic occupancy prediction $\mathbf{O}\in \mathcal{C}^{X \times Y \times Z}$ can be inferred through the probabilistic modeling mechanism:
\vspace{-0mm}
\begin{equation}
     \mathbf{O} = Prob(\mathbf{Q}).
    \label{eq: pred occ}
\end{equation}
For optimization, we adopt the cross entropy loss and the lovaszsoftmax~\cite{lovasz} loss for training.

Due to the lack of structural priors, superquadrics are uniformly initialized in 3D space. 
As a result, some superquadrics in empty regions are optimized to small scales and contribute little to scene modeling, which leads to inefficiency.
To address this, we introduce a pruning-splitting module after initial training. 
Small-scale superquadrics (likely in empty regions) are pruned, while large-scale ones (likely in occupied regions) are split for finer modeling. 
We keep the number of superquadrics unchanged and use two additional blocks to further refine their properties.
Notably, this lightweight module improves superquadric utilization for more efficient scene representation without introducing significant computational overhead.
\section{Experiments}

\label{sec: experiments}
\begin{table*}[t] %
    \caption{\textbf{3D semantic occupancy prediction results on nuScenes.} 
    * means supervised by dense occupancy annotations as opposed to original LiDAR segmentation labels.
    Ch. denotes the channel dimension of our model.
    Our method achieves state-of-the-art performance.}
    \small
    \setlength{\tabcolsep}{0.005\linewidth}  
    \vspace{-2mm}  
    \renewcommand\arraystretch{1.05}
    \centering
    \resizebox{\textwidth}{!}{
    \begin{tabular}{l|c c | c c c c c c c c c c c c c c c c}
        \toprule
        Method
        & IoU
        & mIoU
        & \rotatebox{90}{\textcolor{nbarrier}{$\blacksquare$} barrier}
        & \rotatebox{90}{\textcolor{nbicycle}{$\blacksquare$} bicycle}
        & \rotatebox{90}{\textcolor{nbus}{$\blacksquare$} bus}
        & \rotatebox{90}{\textcolor{ncar}{$\blacksquare$} car}
        & \rotatebox{90}{\textcolor{nconstruct}{$\blacksquare$} const. veh.}
        & \rotatebox{90}{\textcolor{nmotor}{$\blacksquare$} motorcycle}
        & \rotatebox{90}{\textcolor{npedestrian}{$\blacksquare$} pedestrian}
        & \rotatebox{90}{\textcolor{ntraffic}{$\blacksquare$} traffic cone}
        & \rotatebox{90}{\textcolor{ntrailer}{$\blacksquare$} trailer}
        & \rotatebox{90}{\textcolor{ntruck}{$\blacksquare$} truck}
        & \rotatebox{90}{\textcolor{ndriveable}{$\blacksquare$} drive. suf.}
        & \rotatebox{90}{\textcolor{nother}{$\blacksquare$} other flat}
        & \rotatebox{90}{\textcolor{nsidewalk}{$\blacksquare$} sidewalk}
        & \rotatebox{90}{\textcolor{nterrain}{$\blacksquare$} terrain}
        & \rotatebox{90}{\textcolor{nmanmade}{$\blacksquare$} manmade}
        & \rotatebox{90}{\textcolor{nvegetation}{$\blacksquare$} vegetation}
        \\
        \midrule
        MonoScene~\cite{monoscene} & 23.96 & 7.31 & 4.03 &	0.35& 8.00& 8.04&	2.90& 0.28& 1.16&	0.67&	4.01& 4.35&	27.72&	5.20& 15.13&	11.29&	9.03&	14.86 \\
        
        Atlas~\cite{atlas} & 28.66 & 15.00 & 10.64&	5.68&	19.66& 24.94& 8.90&	8.84&	6.47& 3.28&	10.42&	16.21&	34.86&	15.46&	21.89&	20.95&	11.21&	20.54 \\
        
        BEVFormer~\cite{bevformer} & 30.50 & 16.75 & 14.22 &	6.58 & 23.46 & 28.28& 8.66 &10.77& 6.64& 4.05& 11.20&	17.78 & 37.28 & 18.00 & 22.88 & 22.17 & {13.80} &	\textbf{22.21}\\
        
        TPVFormer~\cite{tpvformer} & 11.51 & 11.66 & 16.14&	7.17& 22.63	& 17.13 & 8.83 & 11.39 & 10.46 & 8.23&	9.43 & 17.02 & 8.07 & 13.64 & 13.85 & 10.34 & 4.90 & 7.37\\
        
        TPVFormer*~\cite{tpvformer}  & {30.86} & 17.10 & 15.96&	 5.31& 23.86	& 27.32 & 9.79 & 8.74 & 7.09 & 5.20& 10.97 & 19.22 & {38.87} & {21.25} & {24.26} & {23.15} & 11.73 & 20.81\\

        OccFormer~\cite{occformer} & {31.39} & {19.03} & {18.65} & {10.41} & {23.92} & {30.29} & {10.31} & {14.19} & {13.59} & {10.13} & {12.49} & {20.77} & {38.78} & 19.79 & 24.19 & 22.21 & {13.48} & {21.35}\\
        
        SurroundOcc~\cite{surroundocc} & {31.49} & {20.30}  & {20.59} & {11.68} & \textbf{28.06} & \textbf{30.86} & {10.70} & {15.14} & \textbf{14.09} & \textbf{12.06} & \textbf{14.38} & \textbf{22.26} & 37.29 & {23.70} & {24.49} & {22.77} & \textbf{14.89} & {21.86}  \\

        GaussianFormer~\cite{gaussianformer} & 29.83 & {19.10} & {19.52} & {11.26} & {26.11} & {29.78} & {10.47} & {13.83} & {12.58} & {8.67} & {12.74} & {21.57} & {39.63} & {23.28} & {24.46} & {22.99} & 9.59 & 19.12 \\

        GaussianFormer-2~\cite{huang2024probabilistic} & 30.56 & {20.02} & \textbf{20.15} & {12.99} & {27.61} & {30.23} & {11.19} & {15.31} & {12.64} & {9.63} & {13.31} & \textbf{22.26} & {39.68} & {23.47} & {25.62} & {23.20} & 12.25 & 20.73 \\

        \midrule
        \textbf{QuadricFormer} & \textbf{31.22} & \textbf{20.12} & 19.58 & \textbf{13.11} & 27.27 & 29.64 & \textbf{11.25} & \textbf{16.26} & 12.65 & 9.15 & 12.51 & 21.24 & \textbf{40.20} & \textbf{24.34} & \textbf{25.69} & \textbf{24.24} & 12.95 & 21.86 \\
        
        % \textbf{Ours} (Ch. = 192) & \textbf{31.74} & \textbf{20.82} & \textbf{21.39} & \textbf{13.44} & \textbf{28.49} & 30.82 & 10.92 & \textbf{15.84} & 13.55 & 10.53 & 14.04 & \textbf{22.92} & \textbf{40.61} & \textbf{24.36} & \textbf{26.08} & \textbf{24.27} & 13.83 & 21.98  \\
        
        \bottomrule
    \end{tabular}}
    \label{tab: nuscenes results}
    \vspace{-4mm}
\end{table*}

\begin{table}[t]
    \centering
    \caption{
    \textbf{Performance and efficiency comparison with Gaussian-based methods.}
    The latency and memory are tested on an NVIDIA 4090 GPU with batch size one during inference, in accordance with Gaussian-based methods~\cite{gaussianformer,huang2024probabilistic}.
    Our method achieves better performance-efficiency trade-off.
    }
    \vspace{1mm}
    \setlength{\tabcolsep}{0.02\linewidth}
    \resizebox{1\linewidth}{!}{
    \begin{tabular}{c|ccc|cc}
        \toprule
        Method & {Number of Primitives} & {Latency (ms)} & {Memory (MB)} & mIoU & IoU \\
        \midrule
        \multirow{2}{*}{GaussianFormer~\cite{gaussianformer}} & 25600  & 227 & {4850}  & 16.00  & 28.72 \\
        & 144000 & {372} & 6229  & {19.10}  & {29.83}  \\
        \midrule
        \multirow{4}{*}{\makecell{GaussianFormer-2~\cite{huang2024probabilistic} \\ (Depth Initialized)}} & 1600  & 341 & 3075 & 18.73 & {28.99} \\
        & 3200 & 355 & 3076 & {18.75} & {29.64} \\
        & 6400 & 395 & 3652 & {19.55} & {30.37} \\
        & 12800 & 451 & 4535 & {19.69} & {30.43} \\
        \midrule
        \multirow{4}{*}{\makecell{QuadricFormer \\(Ours)}} & 1600  & \textbf{162} & \textbf{2554} & 20.04 & {30.71} \\
        & 3200 & 164 & {2556} & {20.35} & {31.62} \\
        & 6400 & 165 & 2560 & {20.79} & {31.89} \\
        & 12800 & 179 & 2563 & \textbf{21.11} & \textbf{32.13} \\
        \bottomrule
    \end{tabular}}
    \vspace{-4mm}
    \label{tab:number of gaussians}
\end{table}
\subsection{Datasets and Metrics}
\textbf{NuScenes}~\cite{nuscenes} comprises 1,000 urban driving sequences collected in Boston and Singapore. 
The dataset is officially split into 700 sequences for training, 150 for validation, and 150 for testing. 
Each sequence spans a duration of 20 seconds with RGB images captured by 6 surrounding cameras, and the key frames are annotated at a 2 Hz frequency. 
For supervision and evaluation, we leverage the dense semantic occupancy annotations from SurroundOcc. 
The annotated voxel grid extends from -50m to 50m along both the X and Y axes, and from -5m to 3m along the Z axis, with a spatial resolution of $200\times 200\times 16$. 
Each voxel is classified into one of the 18 categories(16 semantics, 1 empty and 1 unknown).

\textbf{The evaluation metrics} adhere to common practice, namely mean Intersection-over-Union (mIoU) and Intersection-over-Union (IoU):
\begin{equation}
\mathbf{mIoU} = \frac{1}{|\mathcal{C}'|}\sum_{i\in \mathcal{C}'}{\frac{TP_i}{TP_i+FP_i+FN_i}},
\end{equation}
\begin{equation}
\mathbf{IoU} = \frac{TP_{\neq c_0}}{TP_{\neq c_0}+FP_{\neq c_0}+FN_{\neq c_0}},
\end{equation}
Where $\mathcal{C}'$, $c_0$, TP, FP, and FN represent the non-empty classes, the empty class, and the number of true positive, false positive, and false negative predictions, respectively.

\subsection{Implementation Details}
The input images are at resolutions of 900$\times$1600 for nuScenes with random flipping and photometric distortion augmentations. 
We employ ResNet101-DCN with FCOS3D checkpoint for nuScenes.
The numbers of Superquarics are set to 1600 in our main results for nuScenes. 
For optimization, we train our model using AdamW with weight decay of 0.01, and maximum learning rate of $4\times 10^{-4}$, which decays with a cosine schedule. 
We train our model for 20 epochs on nuScenes with a batch of 8.

\subsection{Main Results}
\textbf{3D semantic occupancy prediction.} 
We report the performance of our QuadricFormer in Table~\ref{tab: nuscenes results}.
Compared to other methods, our approach achieves state-of-the-art performance. Specifically, QuadricFormer outperforms other methods on categories such as bicycle, motorcycle, truck and various ground-related classes (drivable surface, sidewalk, terrain, etc.), demonstrating superior capability in modeling both small and structural objects. Moreover, our method significantly surpasses GaussianFormer-2~\cite{huang2024probabilistic} while using substantially fewer superquadrics (1600 vs. 12800), further validating its efficiency and effectiveness.

% Compared to GaussianFormer-2, which employs a significantly larger number of Gaussians and additional initialization strategies, our approach achieves better performance with a more efficient configuration, validating the effectiveness and generalization ability of our method under minimal prior assumptions.

\textbf{Performance and efficiency comparison with Gaussian-based methods.}
We report the performance and efficiency comparison for QuadricFormer with Gaussian-based methods in Table~\ref{tab:number of gaussians}. 
% The results demonstrate a comprehensive comparison between our method (QuadricFormer) and recent Gaussian-based baselines. 
QuadricFormer consistently outperforms prior methods in both 3D semantic occupancy prediction and computational efficiency. 
Specifically, our method achieves the highest mIoU (up to 21.11) and IoU (up to 32.13), surpassing all Gaussian-based approaches.
In terms of efficiency, QuadricFormer significantly reduces both latency and memory usage. For similar or even fewer primitives (e.g., 1600 or 3200), our method achieves a latency as low as 162 ms and 2554 MB memory consumption, which are substantially lower than others.
Notably, even when increasing the number of primitives in QuadricFormer to 12800, both latency and memory usage remain lower than those of Gaussian-based methods using only 1600 primitives. 
This further highlights the superior efficiency of our approach for the complex structures in real-world applications.

% \begin{table}[t]
%     \centering
%     \caption{
%     \textbf{Performance and efficiency comparison with Gaussian-based methods.}
%     The latency and memory are tested on an NVIDIA 4090 GPU with batch size one during inference, in accordance with Gaussian-based methods~\cite{gaussianformer,huang2024probabilistic}.
%     Our method achieves better performance-efficiency trade-off.
%     }
%     \vspace{2mm}
%     \setlength{\tabcolsep}{0.02\linewidth}
%     \resizebox{1\linewidth}{!}{
%     \begin{tabular}{c|ccc|cc}
%         \toprule
%         Method & {Number of Primitives} & {Latency (ms)} & {Memory (MB)} & mIoU & IoU \\
%         \midrule
%         \multirow{2}{*}{GaussianFormer~\cite{gaussianformer}} & 25600  & 227 & {4850}  & 16.00  & 28.72 \\
%         & 144000 & {372} & 6229  & {19.10}  & {29.83}  \\
%         \midrule
%         \multirow{4}{*}{\makecell{GaussianFormer-2~\cite{huang2024probabilistic} \\ (Depth Initialized)}} & 1600  & 341 & 3075 & 18.73 & {28.99} \\
%         & 3200 & 355 & 3076 & {18.75} & {29.64} \\
%         & 6400 & 395 & 3652 & {19.55} & {30.37} \\
%         & 12800 & 451 & 4535 & {19.69} & {30.43} \\
%         \midrule
%         \multirow{4}{*}{\makecell{QuadricFormer \\(Ours)}} & 1600  & \textbf{162} & \textbf{2554} & 20.04 & {30.71} \\
%         & 3200 & 164 & {2556} & {20.35} & {31.62} \\
%         & 6400 & 165 & 2560 & {20.79} & {31.89} \\
%         & 12800 & 179 & 2563 & \textbf{21.11} & \textbf{32.13} \\
%         \bottomrule
%     \end{tabular}}
%     \vspace{-4mm}
%     \label{tab:number of gaussians}
% \end{table}
\begin{figure*}[t]
\centering
\includegraphics[width=1.0\linewidth]{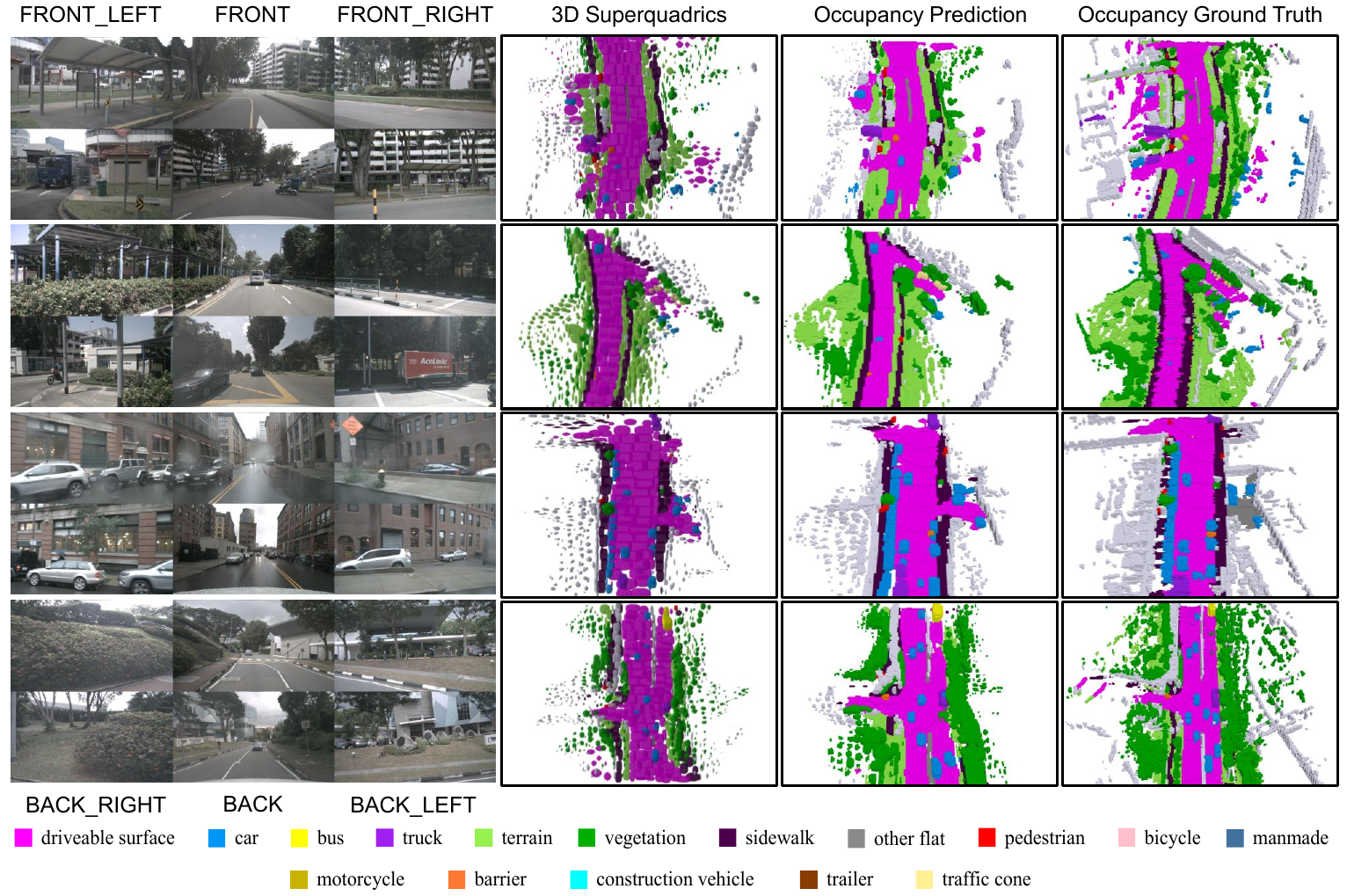}
\vspace{-5mm}
\caption{\textbf{3D Superquadrics and occupancy visualizations on nuScenes.} Our model is able to predict high-fidelity shapes and achieves comprehensive occupancy results.
}
\label{fig:occquadric}
\vspace{-6mm}
\end{figure*}
\subsection{Ablation Study}
% \begin{wraptable}[7]{r}{0.4\textwidth}
% \renewcommand\arraystretch{1.1}
% \centering
% \setlength{\tabcolsep}{12pt}
% \small
% {
% \vspace{-9mm}
% \caption{
% \textbf{Effect of the $\epsilon$ range.}
% \label{tab:range}
% }
% \vspace{3pt}
% % \scalebox{0.98}
% {
% \begin{tabular}{c|cc}
%             \toprule
%             Range of $\epsilon$ & mIoU & IoU \\
%             \midrule
%             $(0.01, 2)$ & 20.39 & 31.13 \\
%             $(0.01, 5)$ & 20.25 & 30.63 \\
%             $(0.1, 2)$ & \textbf{20.51} & \textbf{31.25} \\
%             $(0.1, 5)$ & 19.86 & 30.65 \\
%             \bottomrule
%         \end{tabular}
% }
% \vspace{-8pt}
% }
% \end{wraptable}
% \vspace{1pt}

\textbf{Effect of the $\epsilon$ range.}
We conduct ablation study on the range of the superquadric exponent parameters $\epsilon$ in Eq.~\ref{eq: superquadric}, as reported in Table~\ref{tab:range}. 
We set the number of superquadrics to 12800 for these experiments.
The table explores the effect of different $\epsilon$ ranges on 3D semantic occupancy prediction performance. We observe that setting the range of $(0.1, 2)$ yields the best results, achieving the highest mIoU (20.51) and IoU (31.25).

% We found that when the upper bound of the range exceeds 2 and approaches 5, the quadric shapes tend to become non-convex polyhedra with inward concavities. This results in reduced modeling capability for object shapes. On the other hand, when the lower bound is set below 0.1 but above 0.01, discontinuities may arise on the object boundaries, which also negatively impact geometric modeling. These observations further underline the importance of choosing a moderate range for $\epsilon$ to ensure both shape expressiveness and stability in our representation.

\begin{table}[t]
    \small
    \centering
    \begin{minipage}[t]{0.49\linewidth}
        \centering
        
        \caption{Effect of the $\epsilon$ range.}
        \renewcommand{\tabcolsep}{6mm}
        \begin{tabular}{c|cc}
        \toprule
            Range of $\epsilon$ & mIoU & IoU \\
            \midrule
            $(0.01, 2)$ & 20.39 & 31.13 \\
            $(0.01, 5)$ & 20.25 & 30.63 \\
            $(0.1, 2)$ & \textbf{20.51} & \textbf{31.25} \\
            $(0.1, 5)$ & 19.86 & 30.65 \\
            \bottomrule
        \end{tabular}
        \label{tab:range}
    \end{minipage}
    \hfill
    \begin{minipage}[t]{0.49\linewidth}
        \centering
        \caption{Effect of the pruning-splitting module.}
        \renewcommand{\tabcolsep}{4mm}
        \begin{tabular}{c|cc}
            \toprule
            Crop $\&$ Split Number & mIoU & IoU \\
            \midrule
            0 & 19.41 & 39.77 \\
            200 & 19.65 & 30.35 \\
            400 & 19.90 & 30.67 \\
            800 & \textbf{20.12} & \textbf{31.22} \\
            \bottomrule
        \end{tabular}
        \label{tab:crop}
    \end{minipage}
    \vspace{-3.0mm}
\end{table}

\begin{figure*}[t]
\centering
\includegraphics[width=0.95\linewidth]{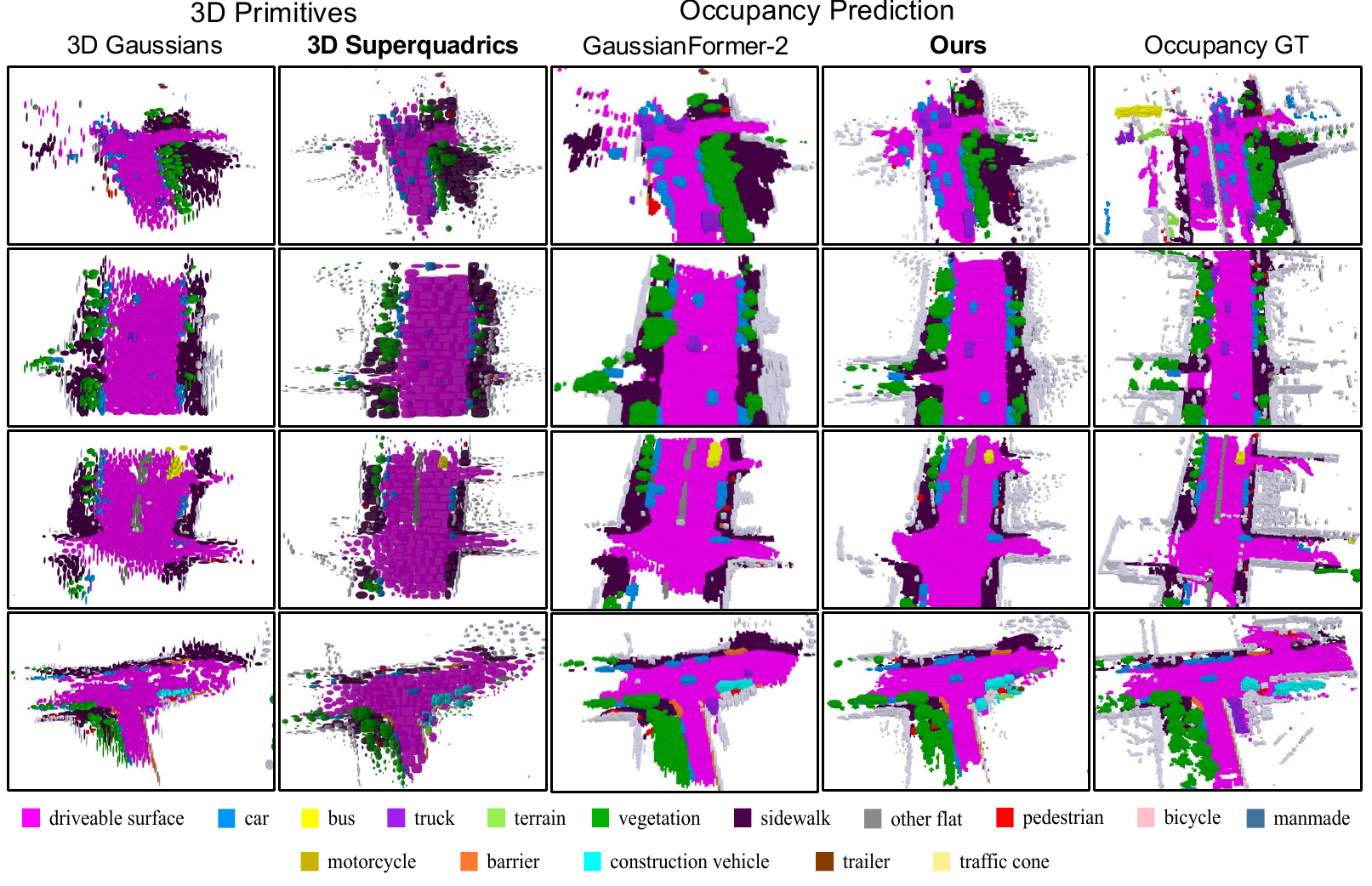}
\vspace{-3mm}
\caption{\textbf{Qualitative comparisons.}
QuadricFormer predicts more flexible and adaptive shapes.
}
\label{fig:comparison}
\vspace{-7.5mm}
\end{figure*}

% \begin{figure*}[t]
% \centering
% \includegraphics[width=0.95\textwidth]{figures/pos.pdf}
% \vspace{-2mm}
% \caption{\textbf{Visualizations of the position learned by different methods.} Our method learns well-structured spatial arrangements with significantly fewer Quadrics.
% }
% \label{fig:pos}
% \vspace{-6mm}
% \end{figure*}
\textbf{Effect of the pruning-splitting module.}
We conduct ablation studies on the effect of the pruning-splitting module, as shown in Table~\ref{tab:crop}. 
% For the experiments, we keep the total number of superquadrics fixed at 6400. The "Crop $\&$ Split Number" denotes how many superquadrics are pruned from low-occupancy (or empty) regions and then redistributed by splitting those in highly occupied areas, thus re-allocating modeling capacity to where it is most needed.
The results demonstrate that increasing the crop $\&$ split number consistently improves performance. This confirms that reallocating primitives from low to high occupancy regions effectively enhances the accuracy and efficiency of our 3D scene representation.

% \begin{table}[t]
%     \centering
%     \caption{
%     \textbf{Ablation on the efficiency of QuadricFormer.}
%     We set the number of Gaussians/Superquarics to 1600.
%     Perc. and Dist. denote the percentage of Gaussians in correct positions, and the average distance of each Gaussian to its nearest occupied voxel, respectively.
%     Overall and Indiv. represent the overall and individual overlapping ratios of Gaussians, respectively.
%     }
%     \vspace{-2mm}
%     \setlength{\tabcolsep}{0.01\linewidth}
%     \resizebox{1\linewidth}{!}{
%     \begin{tabular}{c|cc|cc|cc}
%         \toprule
%         \multirow{2}{*}{Method} & \multicolumn{2}{c|}{Position} & \multicolumn{2}{c|}{Overlap} & \multirow{2}{*}{mIoU}   & \multirow{2}{*}{IoU} \\
%         & Perc. (\%) \( \uparrow \) & Dist. (m) \( \downarrow \) & Overall \( \downarrow \) & Indiv. \( \downarrow \) \\
%         \midrule
%         GaussianFormer~\cite{gaussianformer} & 16.41 & 3.07 & 10.99 & 68.43 & 16.00 & 28.72   \\
%         GaussianFormer-2~\cite{huang2024probabilistic} & 16.41 & 3.07 & 10.99 & 68.43 & 16.00 & 28.72   \\
%         {Ours} & \textbf{28.85} & \textbf{1.24} & \textbf{3.91} & \textbf{12.48} & \textbf{20.32} & \textbf{31.04} \\
%         \bottomrule
%     \end{tabular}}
%     \vspace{-7mm}
%     \label{tab:efficiency}
% \end{table}

\subsection{Visualizations}
We present visualizations of the predicted superquadrics and occupancy results in Figure~\ref{fig:occquadric}.
Our model is able to predict high-fidelity shapes using superquadrics and achieves comprehensive occupancy results.
Further, we compare our method against GaussianFormer-2~\cite{huang2024probabilistic} in Figure~\ref{fig:comparison}, showing that our predicted superquadrics offer more adaptive shapes than Gaussians.
Moreover, our method achieves high-quality performance using only 1600 superquadrics, compared to 6400 Gaussians.
% We also visualize the position learning results using 1600 Quadrics versus 6400 Gaussians in Figure~\ref{fig:pos}, demonstrating that our method learns well-structured spatial arrangements even with significantly fewer Quadrics.

% \begin{figure*}[!ht]
% \centering
% \includegraphics[width=0.95\linewidth]{figures/occ_gaussian.pdf}
% \vspace{-2mm}
% \caption{\textbf{Comparison with GaussianFormer-2.}
% Our method predicts more flexible and adaptive shapes with Quadrics.
% }
% \label{fig:comparison}
% \vspace{-3mm}
% \end{figure*}

% \begin{figure*}[!ht]
% \centering
% \includegraphics[width=0.95\textwidth]{figures/pos.pdf}
% \vspace{-2mm}
% \caption{\textbf{Visualizations of the position learned by different methods.} Our method learns well-structured spatial arrangements with significantly fewer Quadrics.
% }
% \label{fig:pos}
% \vspace{-6mm}
% \end{figure*}
\section{Conclusion}
\label{sec:conclusion}
In this paper, we have proposed a superquadric-based object-centric representation for efficient 3D semantic occupancy prediction.
Specifically, we leverage the geometric expressiveness of superquadrics to model complex structures with far fewer sparsely packed primitives.
We formulate a probabilistic superquadric mixture model, where each superquadric encodes an occupancy probability distribution with a corresponding geometry prior, and semantics are inferred via probabilistic mixture.
Furthermore, we introduce a pruning-and-splitting module that adaptively concentrates superquadrics in occupied regions to further enhance modeling efficiency.
Our proposed QuadricFormer demonstrates state-of-the-art performance and superior efficiency on the nuScenes benchmark.
% , providing an effective and compact solution for scene understanding in vision-centric autonomous driving systems.

\textbf{Limitations.}
With random initialization, QuadricFormer cannot fully learn accurate superquadric positions, leaving some superquadrics in empty regions and reducing representation efficiency.

\textbf{Broader Impact.}
Our work on autonomous driving has the potential to improve traffic efficiency in the future, but it may also contribute to job displacement for drivers.
% \clearpage
% \setcounter{page}{1}
% \maketitlesupplementary
\appendix

{%
\vspace{-10mm}
\begin{center}
    \centering
    \includegraphics[width=\linewidth]{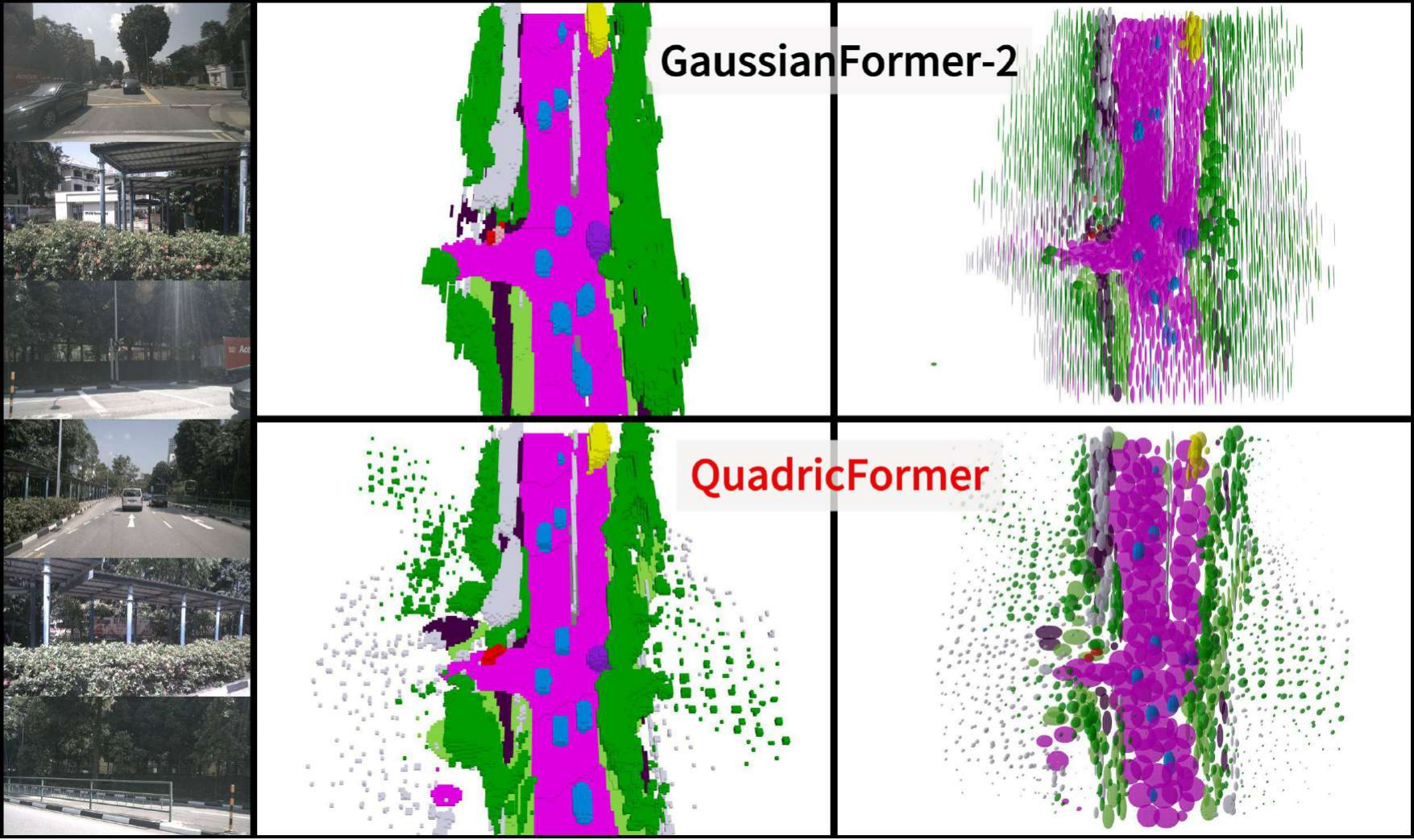}
    \vspace{-4mm}
    \captionof{figure}{
\textbf{Visualizations of the proposed QuadricFormer compared to GaussianFormer-2~\cite{huang2024probabilistic} for 3D semantic occupancy prediction on the nuScenes~\cite{nuscenes} validation set.}
We visualize the six surrounding camera inputs, the corresponding occupancy prediction results, and the primitive representations.
The upper row shows the predicted occupancy(left) and the primitive representation(right) by GaussianFormer-2.
The lower row shows the prediction results of QuadricFormer.
}
\label{video}
\vspace{-3mm}
\end{center}%
}

% \appendix

\section{Additional Superquadric Details}
\begin{figure*}[t]
\centering
\includegraphics[width=0.9\textwidth]{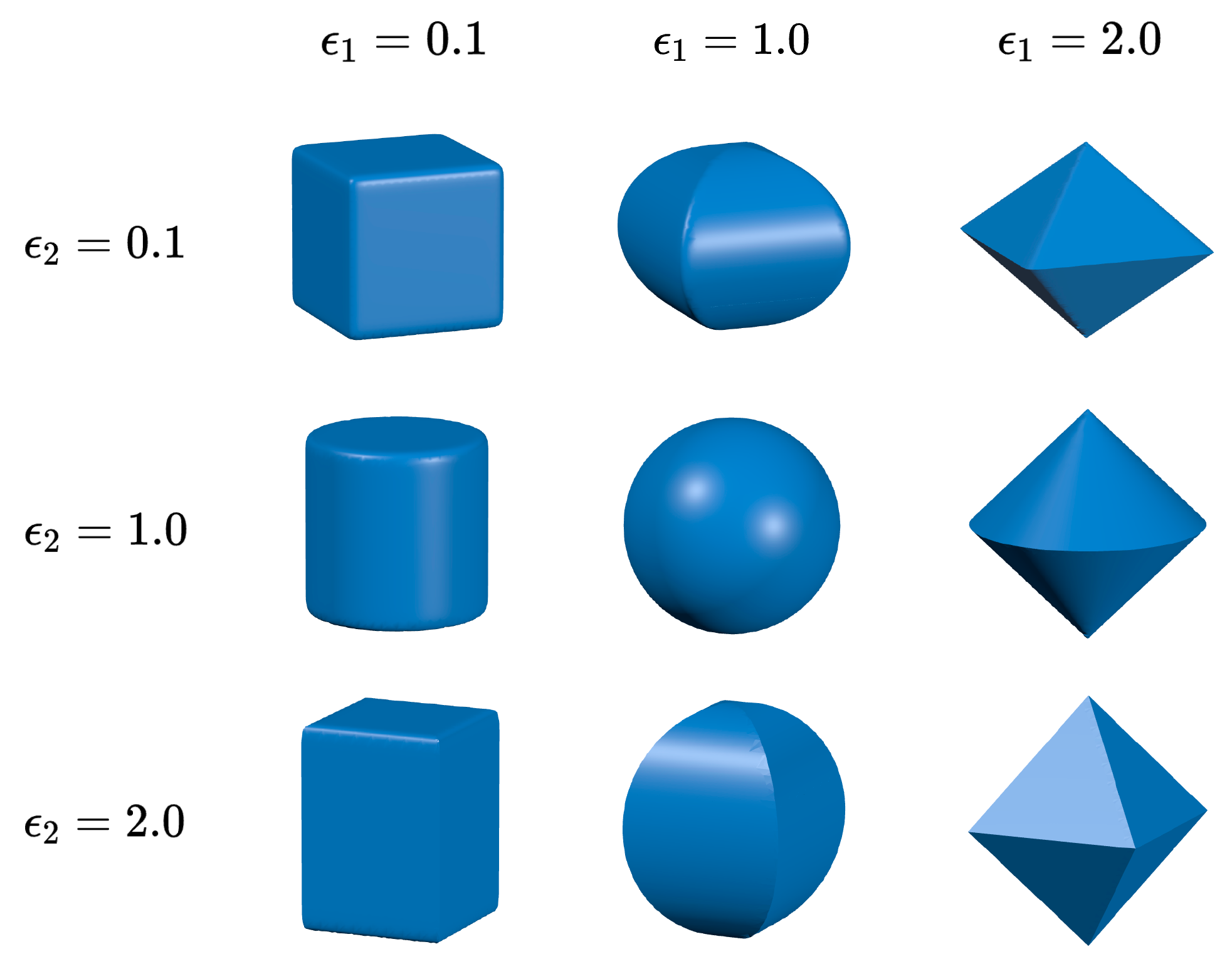}
\vspace{-1mm}
\caption{\textbf{Superquadrics of different shape parameters.} The figure illustrates how varying $\epsilon_1$ and $\epsilon_2$ produces a wide range of shapes, from star-like and rounded shapes to square-like structures. Such diversity enables superquadrics to flexibly model complex object geometries in 3D scenes.
}
\label{fig:quadric}
\vspace{-2mm}
\end{figure*}

Superquadrics are a powerful family of parameterized surfaces that can represent various geometric shapes.
With just a few parameters, superquadrics can generate shapes ranging from basic ellipsoids, cuboids, and cylinders to more complex shapes with rounded corners, star-like profiles, and smooth transitions between them. 
This geometric flexibility makes superquadrics ideal for efficiently modeling diverse objects in autonomous driving scenes.
The shape of a superquadric is mainly controlled by two groups of parameters. 
The first group consists of scaling factors $(s_x,s_y,s_z)$, which define the superquadric's dimensions or "radii" along its three principal axes, determining the object's overall size and aspect ratio.
The second group includes two key shape parameters $(\epsilon_1,\epsilon_2)$ that determine the degree of "squareness" or "roundness" of the object.
$\epsilon_1$ primarily controls the object's profile in planes containing the z-axis (such as the xz- or yz-plane): smaller values (close to 0.1) create sharper profiles, $\epsilon_1=1.0$ produces elliptical outlines, and larger values (up to 2.0) result in flatter contours. 
Similarly, $\epsilon_2$ controls the shape of the cross-section in the xy-plane. A small $\epsilon_2$ yields a star-shaped cross-section, $\epsilon_2$=1.0 gives a circular outline, and large $\epsilon_2$ values produce square-like shapes. 
As shown in Fig~\ref{fig:quadric}, varying $\epsilon_1$ and $\epsilon_2$ of superquadrics results in a wide range of shapes.
By combining these scaling and shape parameters, superquadrics can efficiently represent diverse object geometries in autonomous driving scenes. 
This capability allows them to capture complex structures with significantly fewer primitives than traditional representations (like ellipsoidal Gaussians), highlighting their superior modeling efficiency and expressive power for 3D scene understanding tasks.

\section{Additional Experiments}
\begin{figure*}[h]
\centering
\includegraphics[width=1.0\textwidth]{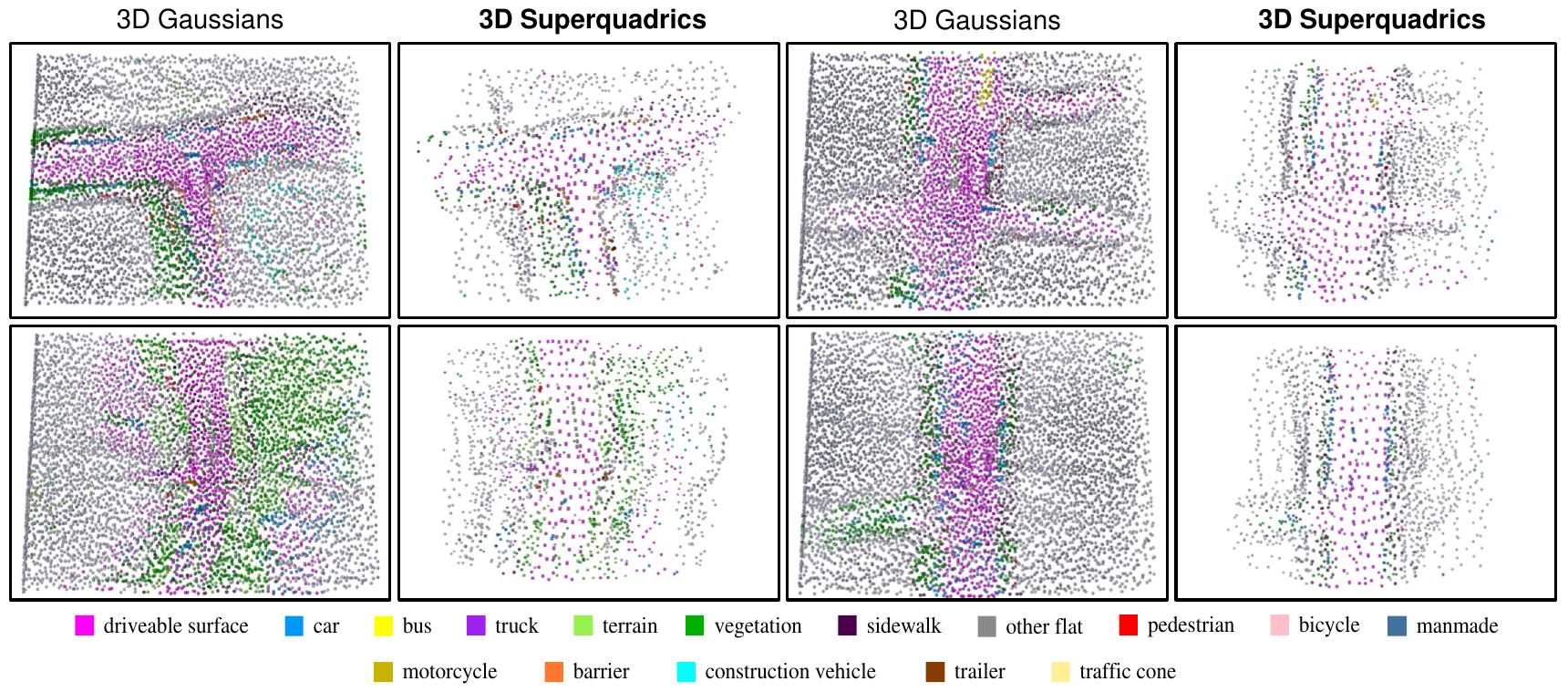}
\vspace{-3mm}
\caption{\textbf{Visualizations of primitive position distributions learned by different methods.} Our approach produces well-structured spatial arrangements while using significantly fewer primitives.
}
\label{fig:pos}
\vspace{-4mm}
\end{figure*}

We visualize the position distributions of scene primitives using 1600 superquadrics versus 6400 Gaussians in Figure~\ref{fig:pos}. 
Gaussian-based methods require a dense arrangement of Gaussians throughout the entire 3D space to model the scene, leading to numerous redundant Gaussians and low modeling efficiency. 
In contrast, our superquadric-based method learns well-structured spatial arrangements, enabling it to effectively model the scene structure with significantly fewer primitives.

\section{Video Demonstration}
Figure~\ref{video} shows a sampled image from the video demo for 3D semantic occupancy prediction on the nuScenes~\cite{nuscenes} validation set.
Compared to GaussianFormer-2~\cite{huang2024probabilistic}, our QuadricFormer exhibits enhanced modeling capability for complex objects and road surfaces.
This demonstrates the effectiveness of our Quadric-based model.

{
%\small
% \bibliographystyle{ieeenat_fullname}
\bibliographystyle{plain}
\bibliography{main}

\begin{thebibliography}{10}

\bibitem{barr1981superquadrics}
Alan~H Barr.
\newblock Superquadrics and angle-preserving transformations.
\newblock {\em IEEE Computer graphics and Applications}, 1(01):11--23, 1981.

\bibitem{lovasz}
Maxim Berman, Amal~Rannen Triki, and Matthew~B Blaschko.
\newblock The lov{\'a}sz-softmax loss: A tractable surrogate for the optimization of the intersection-over-union measure in neural networks.
\newblock In {\em CVPR}, pages 4413--4421, 2018.

\bibitem{nuscenes}
Holger Caesar, Varun Bankiti, Alex~H Lang, Sourabh Vora, Venice~Erin Liong, Qiang Xu, Anush Krishnan, Yu~Pan, Giancarlo Baldan, and Oscar Beijbom.
\newblock nuscenes: A multimodal dataset for autonomous driving.
\newblock In {\em CVPR}, 2020.

\bibitem{monoscene}
Anh-Quan Cao and Raoul de~Charette.
\newblock Monoscene: Monocular 3d semantic scene completion.
\newblock In {\em CVPR}, pages 3991--4001, 2022.

\bibitem{cao2023scenerf}
Anh-Quan Cao and Raoul de~Charette.
\newblock Scenerf: Self-supervised monocular 3d scene reconstruction with radiance fields.
\newblock In {\em ICCV}, pages 9387--9398, 2023.

\bibitem{dsketch}
Xiaokang Chen, Kwan-Yee Lin, Chen Qian, Gang Zeng, and Hongsheng Li.
\newblock 3d sketch-aware semantic scene completion via semi-supervised structure prior.
\newblock In {\em CVPR}, pages 4193--4202, 2020.

\bibitem{MV3D-LiDAR}
Xiaozhi Chen, Huimin Ma, Ji~Wan, Bo~Li, and Tian Xia.
\newblock Multi-view 3d object detection network for autonomous driving.
\newblock In {\em CVPR}, 2017.

\bibitem{chen20193d}
Yueh-Tung Chen, Martin Garbade, and Juergen Gall.
\newblock 3d semantic scene completion from a single depth image using adversarial training.
\newblock In {\em 2019 IEEE International Conference on Image Processing (ICIP)}, pages 1835--1839. IEEE, 2019.

\bibitem{af2s3net}
Ran Cheng, Ryan Razani, Ehsan Taghavi, Enxu Li, and Bingbing Liu.
\newblock 2-s3net: Attentive feature fusion with adaptive feature selection for sparse semantic segmentation network.
\newblock In {\em CVPR}, pages 12547--12556, 2021.

\bibitem{fedele2025superdec}
Elisabetta Fedele, Boyang Sun, Leonidas Guibas, Marc Pollefeys, and Francis Engelmann.
\newblock Superdec: 3d scene decomposition with superquadric primitives.
\newblock {\em arXiv preprint arXiv:2504.00992}, 2025.

\bibitem{hu2023uniAD}
Yihan Hu, Jiazhi Yang, Li~Chen, Keyu Li, Chonghao Sima, Xizhou Zhu, Siqi Chai, Senyao Du, Tianwei Lin, Wenhai Wang, et~al.
\newblock Planning-oriented autonomous driving.
\newblock In {\em CVPR}, pages 17853--17862, 2023.

\bibitem{huang2024probabilistic}
Yuanhui Huang, Amonnut Thammatadatrakoon, Wenzhao Zheng, Yunpeng Zhang, Dalong Du, and Jiwen Lu.
\newblock Probabilistic gaussian superposition for efficient 3d occupancy prediction.
\newblock {\em arXiv preprint arXiv:2412.04384}, 2024.

\bibitem{selfocc}
Yuanhui Huang, Wenzhao Zheng, Borui Zhang, Jie Zhou, and Jiwen Lu.
\newblock Selfocc: Self-supervised vision-based 3d occupancy prediction.
\newblock In {\em CVPR}, pages 19946--19956, 2024.

\bibitem{tpvformer}
Yuanhui Huang, Wenzhao Zheng, Yunpeng Zhang, Jie Zhou, and Jiwen Lu.
\newblock Tri-perspective view for vision-based 3d semantic occupancy prediction.
\newblock In {\em CVPR}, pages 9223--9232, 2023.

\bibitem{gaussianformer}
Yuanhui Huang, Wenzhao Zheng, Yunpeng Zhang, Jie Zhou, and Jiwen Lu.
\newblock Gaussianformer: Scene as gaussians for vision-based 3d semantic occupancy prediction.
\newblock {\em arXiv preprint arXiv:2405.17429}, 2024.

\bibitem{jaklic2000segmentation}
Ales Jaklic, Ales Leonardis, and Franc Solina.
\newblock {\em Segmentation and recovery of superquadrics}, volume~20.
\newblock Springer Science \& Business Media, 2000.

\bibitem{pointpillars}
Alex~H Lang, Sourabh Vora, Holger Caesar, Lubing Zhou, Jiong Yang, and Oscar Beijbom.
\newblock Pointpillars: Fast encoders for object detection from point clouds.
\newblock In {\em CVPR}, 2019.

\bibitem{aicnet}
Jie Li, Kai Han, Peng Wang, Yu~Liu, and Xia Yuan.
\newblock Anisotropic convolutional networks for 3d semantic scene completion.
\newblock In {\em CVPR}, pages 3351--3359, 2020.

\bibitem{voxformer}
Yiming Li, Zhiding Yu, Christopher Choy, Chaowei Xiao, Jose~M Alvarez, Sanja Fidler, Chen Feng, and Anima Anandkumar.
\newblock Voxformer: Sparse voxel transformer for camera-based 3d semantic scene completion.
\newblock In {\em CVPR}, pages 9087--9098, 2023.

\bibitem{bevformer}
Zhiqi Li, Wenhai Wang, Hongyang Li, Enze Xie, Chonghao Sima, Tong Lu, Qiao Yu, and Jifeng Dai.
\newblock Bevformer: Learning bird's-eye-view representation from multi-camera images via spatiotemporal transformers.
\newblock In {\em ECCV}, 2022.

\bibitem{fb-occ}
Zhiqi Li, Zhiding Yu, David Austin, Mingsheng Fang, Shiyi Lan, Jan Kautz, and Jose~M Alvarez.
\newblock Fb-occ: 3d occupancy prediction based on forward-backward view transformation.
\newblock {\em arXiv preprint arXiv:2307.01492}, 2023.

\bibitem{amvnet}
Venice~Erin Liong, Thi Ngoc~Tho Nguyen, Sergi Widjaja, Dhananjai Sharma, and Zhuang~Jie Chong.
\newblock Amvnet: Assertion-based multi-view fusion network for lidar semantic segmentation.
\newblock {\em arXiv preprint arXiv:2012.04934}, 2020.

\bibitem{octreeocc}
Yuhang Lu, Xinge Zhu, Tai Wang, and Yuexin Ma.
\newblock Octreeocc: Efficient and multi-granularity occupancy prediction using octree queries.
\newblock {\em arXiv preprint arXiv:2312.03774}, 2023.

\bibitem{cotr}
Qihang Ma, Xin Tan, Yanyun Qu, Lizhuang Ma, Zhizhong Zhang, and Yuan Xie.
\newblock Cotr: Compact occupancy transformer for vision-based 3d occupancy prediction.
\newblock In {\em CVPR}, pages 19936--19945, 2024.

\bibitem{driveworld}
Chen Min, Dawei Zhao, Liang Xiao, Jian Zhao, Xinli Xu, Zheng Zhu, Lei Jin, Jianshu Li, Yulan Guo, Junliang Xing, et~al.
\newblock Driveworld: 4d pre-trained scene understanding via world models for autonomous driving.
\newblock In {\em CVPR}, pages 15522--15533, 2024.

\bibitem{atlas}
Zak Murez, Tarrence~van As, James Bartolozzi, Ayan Sinha, Vijay Badrinarayanan, and Andrew Rabinovich.
\newblock Atlas: End-to-end 3d scene reconstruction from posed images.
\newblock In {\em ECCV}, pages 414--431, 2020.

\bibitem{lmscnet}
Luis Roldao, Raoul de~Charette, and Anne Verroust-Blondet.
\newblock Lmscnet: Lightweight multiscale 3d semantic completion.
\newblock In {\em 2020 International Conference on 3D Vision (3DV)}, pages 111--119, 2020.

\bibitem{occaspoints}
Yiang Shi, Tianheng Cheng, Qian Zhang, Wenyu Liu, and Xinggang Wang.
\newblock Occupancy as set of points.
\newblock {\em arXiv preprint arXiv:2407.04049}, 2024.

\bibitem{spvnas}
Haotian Tang, Zhijian Liu, Shengyu Zhao, Yujun Lin, Ji~Lin, Hanrui Wang, and Song Han.
\newblock Searching efficient 3d architectures with sparse point-voxel convolution.
\newblock In {\em ECCV}, pages 685--702, 2020.

\bibitem{sparseocc}
Pin Tang, Zhongdao Wang, Guoqing Wang, Jilai Zheng, Xiangxuan Ren, Bailan Feng, and Chao Ma.
\newblock Sparseocc: Rethinking sparse latent representation for vision-based semantic occupancy prediction.
\newblock In {\em CVPR}, pages 15035--15044, 2024.

\bibitem{tian2023occ3d}
Xiaoyu Tian, Tao Jiang, Longfei Yun, Yue Wang, Yilun Wang, and Hang Zhao.
\newblock Occ3d: A large-scale 3d occupancy prediction benchmark for autonomous driving.
\newblock {\em arXiv preprint arXiv:2304.14365}, 2023.

\bibitem{scene_as_occ}
Wenwen Tong, Chonghao Sima, Tai Wang, Li~Chen, Silei Wu, Hanming Deng, Yi~Gu, Lewei Lu, Ping Luo, Dahua Lin, et~al.
\newblock Scene as occupancy.
\newblock In {\em ICCV}, pages 8406--8415, 2023.

\bibitem{wang2024opus}
Jiabao Wang, Zhaojiang Liu, Qiang Meng, Liujiang Yan, Ke~Wang, Jie Yang, Wei Liu, Qibin Hou, and Ming-Ming Cheng.
\newblock Opus: occupancy prediction using a sparse set.
\newblock {\em arXiv preprint arXiv:2409.09350}, 2024.

\bibitem{wang2024occsora}
Lening Wang, Wenzhao Zheng, Yilong Ren, Han Jiang, Zhiyong Cui, Haiyang Yu, and Jiwen Lu.
\newblock Occsora: 4d occupancy generation models as world simulators for autonomous driving.
\newblock {\em arXiv preprint arXiv:2405.20337}, 2024.

\bibitem{openoccupancy}
Xiaofeng Wang, Zheng Zhu, Wenbo Xu, Yunpeng Zhang, Yi~Wei, Xu~Chi, Yun Ye, Dalong Du, Jiwen Lu, and Xingang Wang.
\newblock Openoccupancy: A large scale benchmark for surrounding semantic occupancy perception.
\newblock {\em arXiv preprint arXiv:2303.03991}, 2023.

\bibitem{surroundocc}
Yi~Wei, Linqing Zhao, Wenzhao Zheng, Zheng Zhu, Jie Zhou, and Jiwen Lu.
\newblock Surroundocc: Multi-camera 3d occupancy prediction for autonomous driving.
\newblock In {\em ICCV}, pages 21729--21740, 2023.

\bibitem{js3c}
Xu~Yan, Jiantao Gao, Jie Li, Ruimao Zhang, Zhen Li, Rui Huang, and Shuguang Cui.
\newblock Sparse single sweep lidar point cloud segmentation via learning contextual shape priors from scene completion.
\newblock In {\em AAAI}, volume~35, pages 3101--3109, 2021.

\bibitem{yan2024renderworld}
Ziyang Yan, Wenzhen Dong, Yihua Shao, Yuhang Lu, Liu Haiyang, Jingwen Liu, Haozhe Wang, Zhe Wang, Yan Wang, Fabio Remondino, et~al.
\newblock Renderworld: World model with self-supervised 3d label.
\newblock {\em arXiv preprint arXiv:2409.11356}, 2024.

\bibitem{ye2022lidarmultinet}
Dongqiangzi Ye, Zixiang Zhou, Weijia Chen, Yufei Xie, Yu~Wang, Panqu Wang, and Hassan Foroosh.
\newblock Lidarmultinet: Towards a unified multi-task network for lidar perception.
\newblock {\em arXiv preprint arXiv:2209.09385}, 2022.

\bibitem{drinet++}
Maosheng Ye, Rui Wan, Shuangjie Xu, Tongyi Cao, and Qifeng Chen.
\newblock Drinet++: Efficient voxel-as-point point cloud segmentation.
\newblock {\em arXiv preprint arXiv: 2111.08318}, 2021.

\bibitem{yu2023flashocc}
Zichen Yu, Changyong Shu, Jiajun Deng, Kangjie Lu, Zongdai Liu, Jiangyong Yu, Dawei Yang, Hui Li, and Yan Chen.
\newblock Flashocc: Fast and memory-efficient occupancy prediction via channel-to-height plugin.
\newblock {\em arXiv preprint arXiv:2311.12058}, 2023.

\bibitem{occformer}
Yunpeng Zhang, Zheng Zhu, and Dalong Du.
\newblock Occformer: Dual-path transformer for vision-based 3d semantic occupancy prediction.
\newblock In {\em ICCV}, pages 9433--9443, 2023.

\bibitem{occworld}
Wenzhao Zheng, Weiliang Chen, Yuanhui Huang, Borui Zhang, Yueqi Duan, and Jiwen Lu.
\newblock Occworld: Learning a 3d occupancy world model for autonomous driving.
\newblock In {\em ECCV}, 2024.

\bibitem{zheng2024gaussianad}
Wenzhao Zheng, Junjie Wu, Yao Zheng, Sicheng Zuo, Zixun Xie, Longchao Yang, Yong Pan, Zhihui Hao, Peng Jia, Xianpeng Lang, et~al.
\newblock Gaussianad: Gaussian-centric end-to-end autonomous driving.
\newblock {\em arXiv preprint arXiv:2412.10371}, 2024.

\bibitem{voxelnet}
Yin Zhou and Oncel Tuzel.
\newblock Voxelnet: End-to-end learning for point cloud based 3d object detection.
\newblock In {\em CVPR}, pages 4490--4499, 2018.

\bibitem{pointocc}
Sicheng Zuo, Wenzhao Zheng, Yuanhui Huang, Jie Zhou, and Jiwen Lu.
\newblock Pointocc: Cylindrical tri-perspective view for point-based 3d semantic occupancy prediction.
\newblock {\em arXiv preprint arXiv:2308.16896}, 2023.

\bibitem{zuo2024gaussianworld}
Sicheng Zuo, Wenzhao Zheng, Yuanhui Huang, Jie Zhou, and Jiwen Lu.
\newblock Gaussianworld: Gaussian world model for streaming 3d occupancy prediction.
\newblock {\em arXiv preprint arXiv:2412.10373}, 2024.

\end{thebibliography}
}

\end{document}